\documentclass[pdflatex,sn-apa]{sn-jnl}

\usepackage{graphicx}%
\usepackage{multirow}%
\usepackage{amsmath,amssymb,amsfonts}%
\usepackage{amsthm}%
\usepackage{mathrsfs}%
\usepackage[title]{appendix}%
\usepackage{xcolor}%
\usepackage{textcomp}%
\usepackage{manyfoot}%
\usepackage{booktabs}%
\usepackage{algorithm}%
\usepackage{algorithmicx}%
\usepackage{algpseudocode}%
\usepackage{listings}%
\usepackage{comment}
\usepackage{placeins}
\usepackage{setspace}

\theoremstyle{thmstyleone}%
\theoremstyle{thmstyletwo}%

\theoremstyle{thmstylethree}%

\begin{document}
\title[Article Title]{Human-AI Collaboration: Trade-offs Between Performance and Preferences}


\author*[1]{\fnm{Lukas W.} \sur{Mayer}}\email{lwmayer@uci.edu}
\equalcont{These authors contributed equally to this work.}

\author[1]{\fnm{Sheer} \sur{Karny}}\email{skarny@uci.edu}
\equalcont{These authors contributed equally to this work.}

\author[2]{\fnm{Jackie} \sur{Ayoub}}\email{jackie\_ayoub@honda-ri.com}

\author[2]{\fnm{Miao} \sur{Song}}\email{miao\_song@honda-ri.com}

\author[2]{\fnm{Danyang} \sur{Tian}}\email{danyang\_tian@honda-ri.com}

\author[2]{\fnm{Ehsan} \sur{Moradi-Pari}}\email{emoradipari@honda-ri.com}

\author[1]{\fnm{Mark} \sur{Steyvers}}\email{mark.steyvers@uci.edu}

\affil*[1]{\orgdiv{Department of Cognitive Sciences}, \orgname{University of California, Irvine}, \orgaddress{\city{Irvine}, \state{California}, \country{USA}}}


\affil[2]{\orgname{Honda Research Institute USA, Inc.}, \orgaddress{\city{Ann Arbor}, \state{Michigan}, \country{USA}}}

\abstract{
Despite the growing interest in collaborative AI, designing systems that seamlessly integrate human input remains a major challenge. In this study, we developed a task to systematically examine human preferences for collaborative agents. We created and evaluated five collaborative AI agents with strategies that differ in the manner and degree they adapt to human actions. Participants interacted with a subset of these agents, evaluated their perceived traits, and selected their preferred agent. We used a Bayesian model to understand how agents' strategies influence the Human-AI team performance, AI's perceived traits, and the factors shaping human-preferences in pairwise agent comparisons. Our results show that agents who are more considerate of human actions are preferred over purely performance-maximizing agents. Moreover, we show that such human-centric design can improve the likability of AI collaborators without reducing performance. We find evidence for inequality-aversion effects being a driver of human choices, suggesting that people prefer collaborative agents which allow them to meaningfully contribute to the team. Taken together, these findings demonstrate how collaboration with AI can benefit from development efforts which include both subjective and objective metrics.
}


\maketitle

\section*{Significance}
 Human-AI collaboration is expected to grow in the coming years. Particular attention is being paid to agentic cooperative AI that is capable of autonomously performing helpful tasks without repeated human instruction due to its potential to significantly improve the performance of human-AI teams. However, the use of cooperative AI agents poses two key challenges: (1) the development of such agents in modern multi-agent reinforcement learning paradigms often excludes  human collaborators, and (2) the process of integrating human preferences into the algorithms underlying AI agents remains poorly understood.  Our study addresses these shortcomings by establishing an empirical framework to evaluate how algorithmic changes can be mapped to human preferences. Our study reveals key dynamics, such as algorithm changes that increase human liking of the AI agent without harming the performance of the human-AI team, and a pronounced human preference for inequity-aversion. These findings inform human-AI development by demonstrating how collaborative AI can be both effective and enjoyable. Our approach adjusts agent behavior by modifying algorithmic  inputs and outputs, making it broadly applicable to new and existing agentic systems.

\newpage

\section*{Introduction}\label{sec1}
Contemporary AI technologies have matured to the point where their integration into everyday activities has become feasible. This integration is taking place in a wide range of fields including healthcare, education, and gaming \citep{aiindexreport}. One setting for AI integration that is becoming increasingly common involves a user prompting an AI, for example a chatbot. Here, the user explicitly instructs the AI to produce an output or the AI simply offers non-binding suggestions to the user \citep{Bansal2019ACF, Vodrahalli2022UncalibratedMC}. In these types of interactions human-AI collaboration happens sequentially: A user prompt is followed by an AI response with the human always remaining the ultimate decision-maker. One alternative setting that has gained interest recently is collaboration with an agentic AI, where an AI agent can take actions independently from the human \citep{carroll2019utility, crandall2018cooperating, mckee2024warmth, strouse2021collaborating, bennett2022human, puig2023habitat, nalepka2019human}. An AI being able to act independently from the human could purportedly yield productivity gains due to the ability to concurrently distribute labor across both the human and this agentic AI. For example, imagine that a human and an AI are collaborating on a software project with multiple outstanding tasks. Instead of merely advising the human on how to tackle each task, as in the chatbot setting, the agent might independently address some of the more routine tasks, thus leaving more time for the human to focus on the more challenging tasks. 

Current machine learning research typically views agentic AI designed to collaborate with people as a multi-agent reinforcement learning (MARL) problem \citep{gronauerMultiagentDeepReinforcement2022}. A common difficulty in MARL research is  integrating human considerations into the algorithms that give rise to the agentic AI's behavior. For example, researchers have modeled agents from human demonstrations (behavior cloning), but this method has several notable limitations. First, human data is relatively costly to collect. Second, training an algorithm to reproduce human-like behavior does not explicitly integrate validated design principles \citep{codevilla2019exploring}. Finally, behavior cloning struggles to perform as well as more simulation-based methods \citep{strouse2021collaborating}. This combination of factors limits the applicability of behavior-cloning approaches.

Due to the complexity of integrating humans in the modeling, the development of human-in-the-loop multi-agent systems often neglects recent behavioral studies, which have shown that effective collaborative AI should take into account subjective factors beyond objective performance measures \citep{crandall2018cooperating, ho2016feature, mckee2024warmth, puig2020watch, siu2021evaluation, tang2022exploring,zhang2021ideal, zhang2023understanding}. For example, people prefer an AI agent whose behavior is predictable and transparent, as these characteristics make the AI's actions more understandable and reliable \citep{crandall2018cooperating, ho2016feature, tang2022exploring, zhang2021ideal, zhang2023understanding}. 
Similarly, people prefer non-adaptive, rule-based agents over learning-based agents due to their predictability and ease of interaction \citep{siu2021evaluation}.
The contradiction is evident when we compare human preferences like predictability and simplicity to a contemporary MARL algorithm, which involves complex and adaptive behaviors that are inherently opaque in their decision-making \citep{gronauerMultiagentDeepReinforcement2022}. 
Finally, there is evidence to suggest that people expect collaborative AI agents to exhibit certain behavioral characteristics, or ``traits''. These traits include behaviors that elicit perceptions of warmth, competence, intentionality, and fun \citep{mckee2024warmth, siu2021evaluation}; concepts that receive little attention in algorithm development. Given this divide between the technical development of agentic AI systems and people's expectations for a collaborative agent, there is a growing  need for paradigms that can shape algorithm development in a manner that is compatible with human preferences. 

Addressing human preferences when developing collaborative AI agents is of significant concern since the adoption of AI agents will critically depend on human users' acceptance \citep{SteyversKumar}, yet research  concerned with the development of collaborative AI is generally focused on either multi-agent performance metrics or human-centered designs, rarely both \citep{bhambri2023benchmarking, mckee2024warmth}. In this study, we aim to show an avenue for building effective, human-aligned, collaborative AI by combining performance-driven AI designs and human-centered design approaches. Specifically, we evaluate the impact of distinct AI collaboration strategies on performance and human users' perceptions. To do this, we developed a set of rule-based agents, each representing variations of an egocentric, performance-maximizing agent that incorporates additional manipulations meant to reflect one or more behavioral traits. In addition, we conducted two behavioral experiments that systematically evaluate how these algorithmic variations impact both the performance of the human-AI team and the perceived traits of the AI from the human's perspective. 

Our study addresses two central questions. First, which factors most influence human preferences for collaborative AI agents? In our experiments, we examined both objective metrics related to performance as well as subjective factors related to people's perception of the AI agents' various behavioral traits. Second, do the trade-offs between AI performance and human preference always operate in a strictly linear zero-sum manner where improving one inherently detracts from the other? Alternatively, are there strategies or design choices that can achieve a net positive effect—where a marginal decrease in AI performance (if any) is more than compensated by significant improvements in human performance and approval?

\begin{figure}
    \centering
    \includegraphics[width=0.8\textwidth]{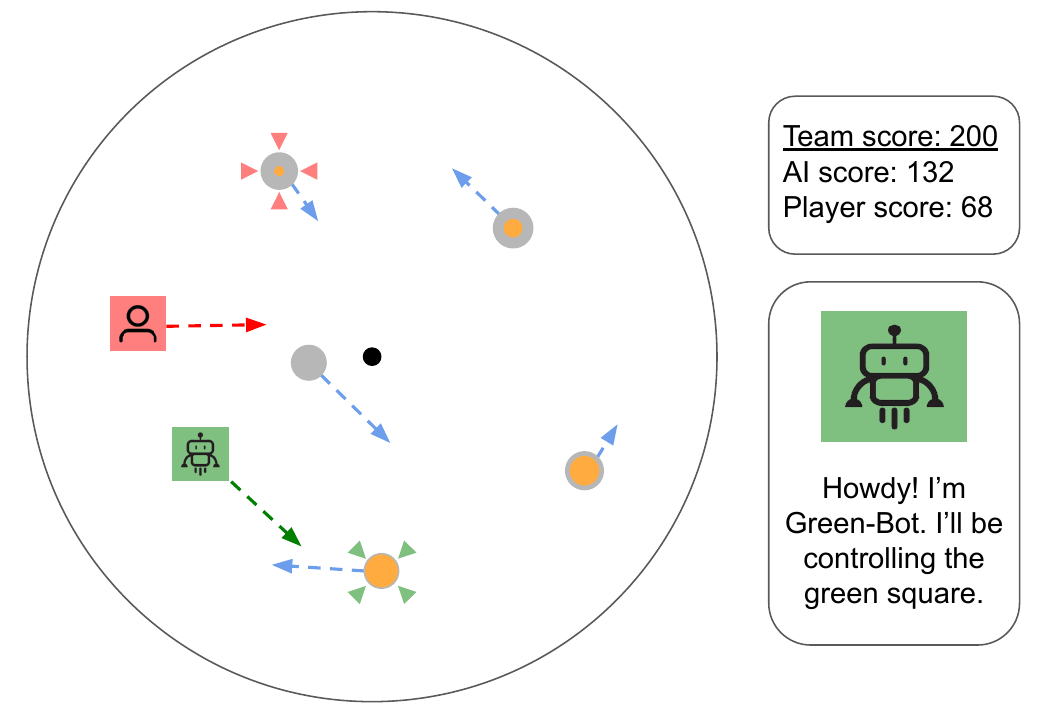}
    \caption{Illustration of the collaborative target interception task with a human player and an AI agent. The game is played in a circular environment where the participant (red avatar) and the AI agent (green avatar) have to collect points by intercepting moving targets (circles) that appear in the game area. New targets appear in the game area, move along a straight path, and then disappear again once they reach the game's edge. Players can click on a target to direct their avatar to the optimal interception point. Arrows are used to illustrate the path and speed of motion of targets and players, but they do not appear in the game environment. The cross-hairs on the targets indicate which target each agent is pursuing. Targets have different point values, as indicated by the orange fill. The game displays score metrics for both individual players and the team (right). Participants interact with various AI agents represented by color names.}
    \label{fig:taskEnvironment}
\end{figure}

\subsection*{Behavioral Experiments}
 To address these questions, we conducted two behavioral experiments to systematically evaluate various AI agent designs, examining how different collaborative strategies affect both performance and human preferences. In our study, human participants interacted with several variations of a collaborative AI agent within a dynamic decision-making task. By analyzing participants' experiences and preferences when collaborating with different types of agents, we seek to identify the key factors that contribute to effective AI collaboration. Importantly, we provide a paradigm in which different algorithmic manipulations can be mapped onto subjective perceptions, allowing us to evaluate whether our manipulations approximate commonly reported design principles. Thus, our experiments provide an approach for bridging the gap between performance-oriented algorithm design and user-centered preferences in human-AI collaboration.

\subsubsection*{Collaborative Target Interception Task}
To investigate human preferences for collaborative AI agents, we created a task in which a human player and an AI agent work together towards a common goal: \textit{achieving a maximally high team score}. Our task is an extension of a dynamic decision-making task previously used to study how humans adopt AI assistance \citep{karny2024learning}. The objective of the game is to collect as many points as possible by intercepting point-valued targets which move at constant speeds through the game environment (see Figure \ref{fig:taskEnvironment}). The task has some of the planning requirements of traveling salesman problems that have been studied in the context of human problem solving \citep{graham2000traveling}, although it additionally involves moving destinations. Importantly, the task necessitates collaborative planning between the human player and the AI agent, as targets vanish after being intercepted by a player or exiting the game view. New targets appear at random intervals, meaning inconsiderate collaborators can ultimately get in each other's way. Each player clicks on targets to direct their avatar to the best interception point. This means that the interface handles the navigation while player's focus on the decision-making. 

A key aspect of the task is the \textit{target density}, which dictates the number of targets that can be present simultaneously in the game environment. By changing the target density, players face different collaborative demands. When many targets are available, both the human player and the AI agent have many targets to choose from, rarely resulting in redundant pursuits. However, when there are few targets, the human player and the AI agent must avoid following the same target to maximize team performance. Effective delegation also ensures that each player does not miss opportunities to intercept other valuable targets before they become unavailable.

\subsubsection*{Collaborative Agents}
Participants were assigned two out of the five agents we developed. They played one round of our experimental task with each of these two agents, before evaluating these AI collaborators in a Likert questionnaire, indicating their preferred agent\footnote{This ``forced-choice'' is the foundation for human preference in this study.} and writing open-ended statements that justified their choices. This procedure was repeated once, so that participants experienced the agents in both target density settings (Figure \ref{fig:within-block} for an overview of the experimental procedure).  

Each of the five AI agents is a variant of a planning algorithm with additional constraints that modify their behavior. These additional constraints reflect different rules that could be thought of as promoting collaborative behaviors. Examples include the avoidance of interfering with ongoing target interceptions, seeking spatial separation to minimize overlap with human actions, mimicking human decision-making capabilities, and focusing on targets the participant is otherwise unlikely to pursue. For details on the collaborative strategies, see the Methods section.  

\section*{Experiment 1}
\subsection*{Methods}\label{sec11}
\subsubsection*{Participants}
300 participants were recruited from the online participant recruitment platform Prolific \citep{prolific}. 287 of these 300 participants were included in the analyses, with the remaining 13 excluded for having incomplete responses. Ages ranged from 18 to 84 (Mean = 35.3, SD = 12.4), with 53\% identifying as female, 46\% as male, and 1\% abstaining from gender identification. All participants were residents of the United States and had not taken part in any of our previous experiments. The study was conducted on participants' personal computers, and each participant was compensated with 5 USD for their participation in the 25-minute experiment. The average compensation rate was 13 USD per hour.

Informed consent was obtained from each participant before the study commenced. The study protocols were approved by the Institutional Review Board of the University of California, Irvine (IRB \#4527), and the study was conducted in accordance with the principles of the Declaration of Helsinki. Participants were assured of the confidentiality of their responses and informed of their right to withdraw from the study at any time without penalty.

\subsubsection*{Game Environment}
There are two agents in the game: a collaborative AI player and a human player, each of whom have their own unique icon and distinctively colored square. The  objective of the game is to intercept as many moving, point-valued targets as possible within a fixed time frame. Both the human player and the AI player can intercept targets but each target can be intercepted only once. Optimal task performance necessitates quick strategic decisions from the human player to intercept targets in a particular sequence during the limited time they are available while also paying attention to and coordinating with the actions of the collaborative AI player. 

Targets spawn randomly at the edge of the circular game area. Their initial movement angle is randomly set within a cone of possible angles. Spawned targets move in straight-line directions at constant speed, sampled from a uniform distribution that ensures target speeds are between 1-50\% slower than the player's avatar. Targets exit the playable area if they are not intercepted. The spawning process ensures that the number of objects in the game area is limited to the target density (either 5 or 15 targets). One key feature of the spawning process is that it is independent of the player's skill in interception. After a player intercepts a target, it disappears from the game area, but its path is still computed until it hits the perimeter. Only once a target, visible or not, hits this perimeter will a new target be spawned. Each target is worth between zero and fifteen points, with the probability distribution of point values following a Beta(1,2) distribution that we discretized over point values via binning. In practice, this means that low-value targets appear more often than high-value targets. 

Players click on targets in order to intercept them. A target click initiates an interception algorithm to calculate the optimal interception path for the player's avatar \footnote{The interception algorithm is based on a time-constrained quadratic equation.}. 
At any point in time, the player can click on different targets to change the path of interception, meaning current trajectories can be interrupted. There is also the option of clicking on a shaded point in the center that allows players to traverse back to the center of the play area.
It is not guaranteed that a player will intercept the target once clicked. The interception point can lie outside the playable game area if the target is too far away for the player to intercept in time. As a result, the player's avatar is guided to the edge of the playable area. The player's avatar will not re-navigate automatically and thus will remain at the edge of the map until the player chooses a new navigation target. 
Note that players can intercept targets that are not explicitly chosen for interception. That is, if a target lies on the path of interception to the chosen target, it will also be intercepted, and its point value will be added to the total.
A colored cross-hair made of four triangles highlights the target currently being pursued. The target marker's color is congruent with the agent's identifying color and indicates the current target each player is pursuing. Both agents can have active cross-hairs on the same target without visual overlap as the AI player's cross-hairs are rotated by 45 degrees.

The game's user interface also includes a display indicating the team score, player score, and AI score. Adjacent to the game area is information to support the player in keeping track of the AI player identities. Here, the icon of the collaborative AI is displayed along with a message that identifies its appearance in the game (e.g., ``Howdy! I'm Green-Bot. I'll be controlling the green square.'' See Figure \ref{fig:taskEnvironment} for an illustration of the interface.

\subsubsection*{Collaborative AI Agents}
We designed five different AI agents to collaborate with human players in the target interception task. Each of the five agents is a variant of a basic search algorithm capable of planning target interception sequences with up to three targets \citep{karny2024learning}. These modifications aimed to improve the interaction between the AI and human players by addressing specific challenges to collaboration in our task.

\paragraph{Search Algorithm} The search algorithm is designed to approximate optimal solutions to the target interception task. The algorithm computes all possible interception sequences involving up to three targets, updating the positions of both the AI player and the targets throughout the sequence. This ensures that the AI player can respond to dynamic changes in the game state. The three-target limit is imposed to ensure the algorithm can operate in real time during behavioral experiments. For more details, see Appendix \ref{searchalgorithm}.

\begin{figure}
    \centering
    \includegraphics[width=1.0\textwidth]{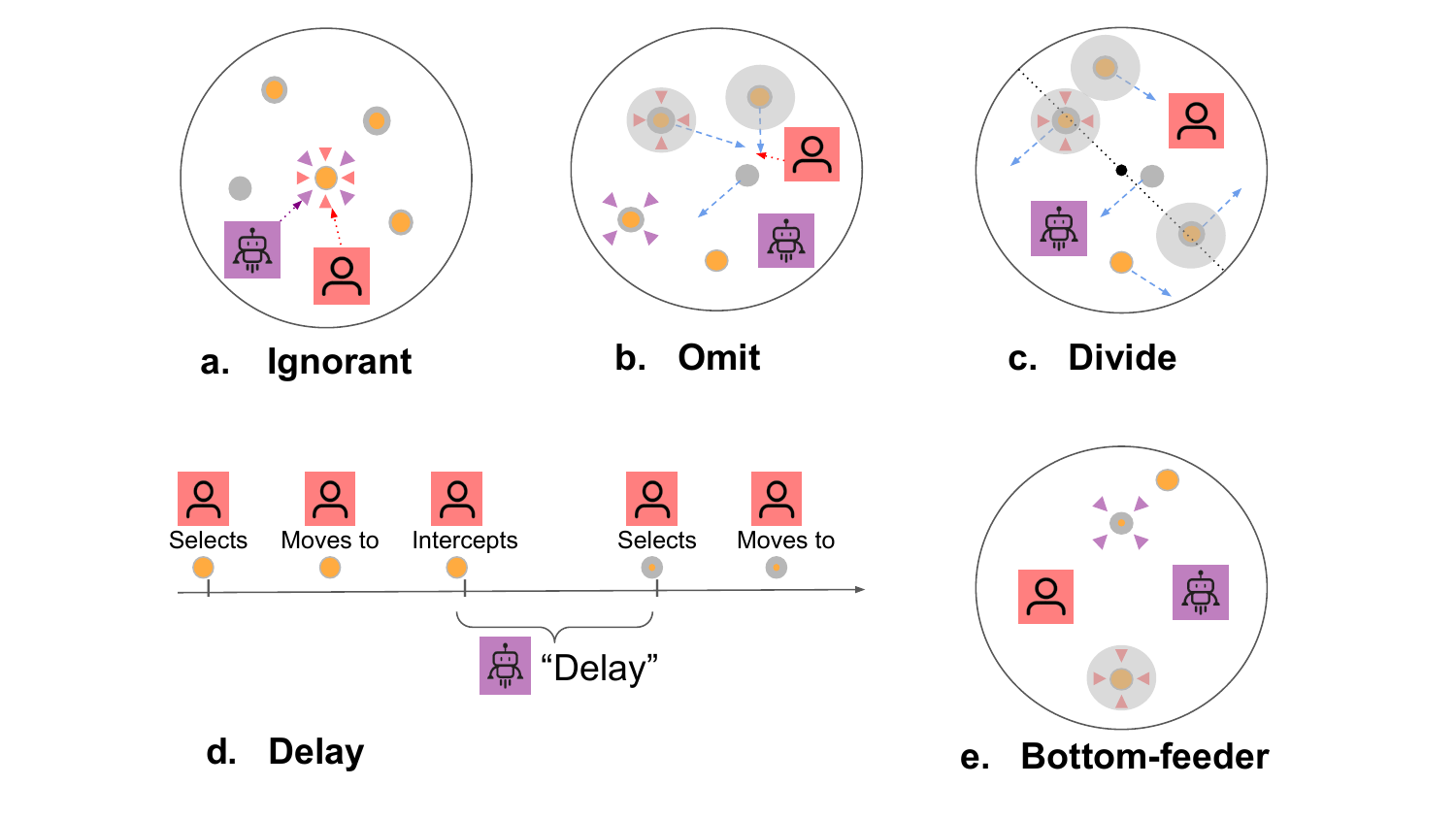}
    \caption{Overview of AI collaborator behavior differences.  \textbf{a.} The Ignorant agent always pursues the highest value target, no matter what the human does. \textbf{b.} The Omit agent ``omits'' targets that the human is intended or predicted to intercept from consideration and is equivalent to Ignorant otherwise. \textbf{c.} The Divide agent extends the logic of Omit by also only considering targets on its half of the game environment. \textbf{d.} The Delay agent approximates the reaction time the human is demonstrating and is otherwise equivalent to Omit. \textbf{e.} The Bottom-Feeder ``inverts'' the value function of Omit, so it always pursues the lowest value target. Note that grayed out targets are not visible to the search algorithm.  }
    \label{fig:agentsoverview}
\end{figure}

\paragraph{Agent Variations} The search algorithm formed the basis for all agents in our experiment. We developed several variations of the search algorithm to create different collaborative AI players. These variations included changes to the target consideration set (which targets the AI player could pursue), delays in initiating a plan, and perception of point values. Variations were conceived as mechanisms that incorporate and give rise to heightened perceptions of traits observed in previous research \citep{mckee2024warmth, siu2021evaluation}. For a graphical overview of agent types see Figure \ref{fig:agentsoverview}. This study is not designed to test all possible combinations of these features. Instead, we focused on a set of five agents that test out a key set of variations: 

\begin{enumerate}
    \item \textbf{Ignorant Agent:}  Our baseline agent uses a basic search algorithm to plan an optimal interception sequence over three objects currently in the game environment. The agent is not provided with information about human intent — it is \textit{ignorant} about which target the human has clicked and is in the process of intercepting. Therefore, the ignorant agent can pursue the same target as the human. Overall, the agent is egocentric in that it does not change its behavior in response to the human player's actions. This agent serves as a baseline comparison for the other agents, as this agent is the least considerate of the human's actions.
    \item \textbf{Omit Agent:} This agent operates with the same search algorithm as the Ignorant agent. However, it is provided with information about the human intended target and can reason about the set of other targets that the human will intercept on its way to the intended target. The agent \textit{omits} this set of targets from the consideration set of targets, meaning these targets cannot be part of the agent's interception plan. If the human player clicks on a new target, the consideration set will be recalculated, so that targets previously clicked by the human are included, but the new currently marked target by the human is not. Dynamic updating also applies to the targets that will be intercepted by the human if they complete their current path. The next three agents are all variations of the omit agent.   
    \item \textbf{Divide Agent:} This agent operates in the same fashion as the omit agent but applies a \textit{divide and conquer} strategy. This was done to make it easier for the human player and AI agent to avoid getting in each other's way and, potentially, have better task delegation \citep{wu2021too, bennett2022human}. The agent (virtually) divides the game area into two halves where the dividing line is orthogonal to the imaginary line from the human player to the game's center-point. The agent only considers targets that can be intercepted in the half not occupied by the human player, with the allotted area being continuously recomputed as the human's position changes. Therefore, this agent omits a larger set of targets from consideration than the omit agent, leaving more targets for the human player, further reducing potential interception conflicts between the players. 
    \item \textbf{Delay Agent:} This agent operates in the same fashion as the omit agent but adds a \textit{delay} between the time a target is intercepted and the selection of a new target to pursue. This artificial delay is designed to decrease the difference in performance relative to the human, as the AI no longer reaches superior performance merely by executing actions more quickly than the human player. This delay is set to adaptively approximate the human player's response times (RTs) throughout the experiment with an exponential moving average of the previous five response times, where a response time constitutes the stretch of time between the point at which an ongoing action is completed and the point at which a new action is initiated.    
    \item \textbf{Bottom-Feeder Agent:} This agent is based on the omit agent but makes changes to the objective function by inverting the target values. Like an ecological \textit{bottom-feeder} that consumes lower-value resources, this agent will consistently target the lowest-value targets available to it. This reduces competition between the human and AI for valuable resources, allowing the human to focus on intercepting the most rewarding targets. While this strategy appears irrational at face value, it may serve collaboration by ensuring that human and AI actions complement each other.
\end{enumerate}

\subsubsection*{Procedure}
Participants accessed the study via Prolific and began by completing a consent form. They then went through an interactive tutorial that explained the game's mechanics. Before commencing with the main experiment, participants were required to demonstrate an understanding of these game mechanics. Thus, participants were informed that the premise of this study is to evaluate how people play alongside a collaborative AI robot in quickly changing environments. 

\begin{figure}
    \centering
    \includegraphics[width=1\textwidth]{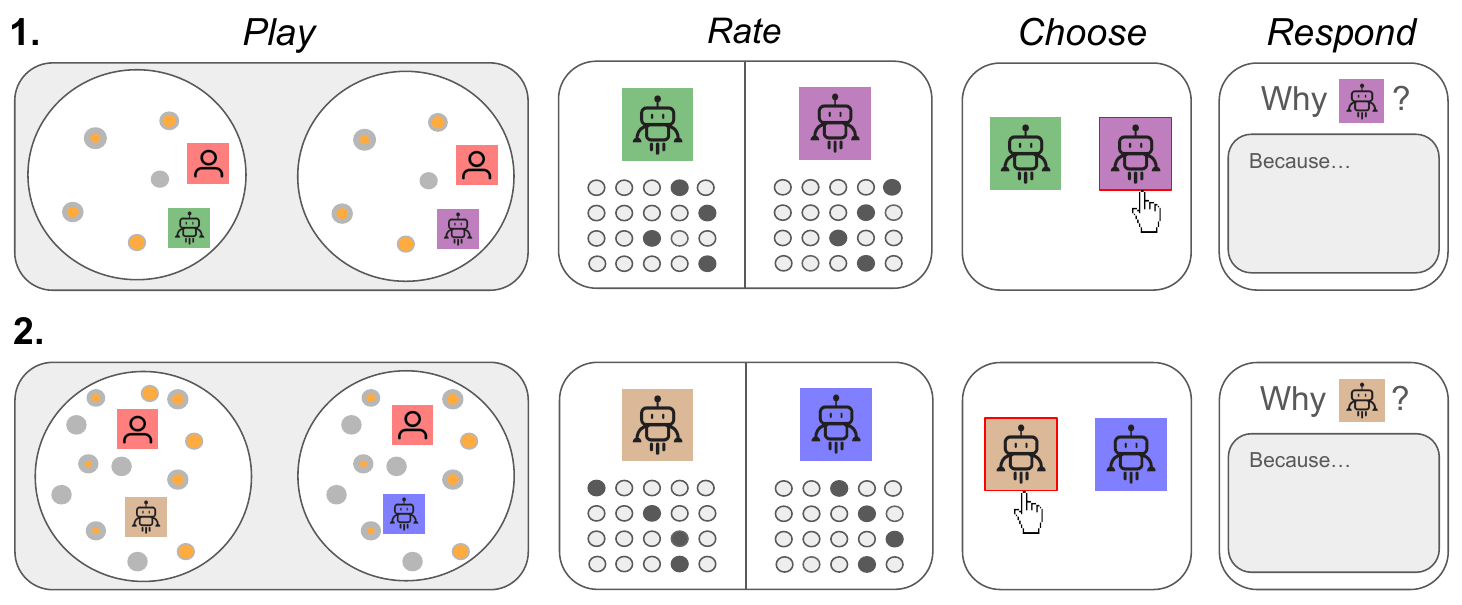}
    \caption{Illustration of the procedure in Experiment 1. Top and bottom rows (\textbf{1} and \textbf{2}) illustrate the two blocks in the experiment. Within each block, participants play two rounds, one with each AI agent from their assigned pair. Agents are then evaluated on a variety of dimensions using 7-point Likert scales. After submitting their ratings, participants indicate their preferred agent in a two-alternative forced choice. Finally, participants are asked to provide free-text responses explaining why they chose the agent they preferred. This procedure is repeated over two blocks where target density is varied. In the illustration, the first and second blocks have low and high target densities respectively. The density order is counterbalanced in the experiment.}
    \label{fig:within-block}
\end{figure}

In the main experiment, participants played two blocks, each with two 3-minute rounds (see Figure \ref{fig:within-block}). Each block featured a different target density (5 or 15). Participants played one round with each of the two collaborative agents per block.

Even though participants played with the same pair of agents in the first and second blocks (at different target densities), this information was not made explicit to participants. In fact, participants would have had reasons to believe that the bots they experienced in the first block are different from those in the second block. For example, participants were informed that they were playing with the green and purple bots in rounds 1 and 2, and the copper and blue bots in rounds 3 and 4. Each bot had two color variations, disguising the fact that participants interacted with only two agents. This identity distinction was compounded by the change in target density, making it relatively hard to compare the behavior of the agents in the first block with those featured in the second block. Obscuring the identity of the AI collaborator through these measures served to ensure that participant judgments were made independently from previous rounds. 

At the end of each block, participants rated each agent on eight dimensions of collaborative ability and performance. Table \ref{tab:survey-summary} shows the list of survey questions, which were based in part on prior research \citep{attigHanabi2024, siu2021evaluation}. The questionnaire was presented in a matrix format, with one matrix for each agent. Each row of the matrix contained a question item with Likert-scale values as the columns. The two matrices were placed next to each other so that the left-hand matrix pertained to one agent with the right-hand matrix corresponding to the other agent. After rating the pair of agents, participants were presented with a choice screen where they selected the agent they preferred to play with. Upon making their selection, participants were prompted to provide an open-ended response explaining their choice. The interface enforced a minimum character limit, requiring participants to write at least a few words. At the conclusion of the study, participants were given the opportunity to provide general open-ended feedback.

\subsubsection*{Design}
The study followed a mixed within- and between-subjects design. The target density was a within-subjects variable, since each participant performed the task with 5 and 15 maximum concurrent targets. The between-subjects variable included the assignment of AI agents to participants. Each participant was assigned two of the five collaborative AI agents. Each participant played a round with each of the two agents for each target density level. The ordering of agents across rounds and the ordering of target densities across blocks was counterbalanced. 

\subsubsection*{Data Analysis}
To assess statistical significance, we utilized Bayes factors (\emph{BF}s) to determine the extent to which the observed data adjust the a priori belief in the alternative and null hypotheses. Values of 3 $<$ \emph{BF} $<$ 10 and \emph{BF} $>$ 10 indicate moderate and strong evidence against the null hypothesis, respectively. Similarly, values of 1/10 $<$ \emph{BF} $<$ 1/3 and \emph{BF} $<$ 1/10 indicate moderate and strong evidence in favor of the null hypothesis, respectively \citep{Jefferys1961, Rouder2009, Rouder2012}. 

The analysis of performance and the questionnaire scores was performed using Bayesian ANOVAs and follow-up T-tests. All statistical results related to Bayes factors were implemented with the BayesFactor package (Version: 0.9.12-4.7) in the R statistical computing software \citep{bfpackage}. Since we performed sensitivity analyses for our Bayesian inferential statistics, the main paper only reports key, prior-robust results in the interest of brevity. The full set of results with sensitivity analyses and code are openly accessible at  \href{https://osf.io/ybweq/?view_only=cb4d4c7ac0b848b79b6ae8c7b09278cc}{this project's OSF page}.

Estimation of the logistic regression in Equation \ref{eq:choice} was performed with Bayesian methods in the JASP (Version 0.19, \cite{JASP2024}) environment using the default priors based on the Generalized g-Prior Distribution (CCH; \cite{li2018mixtures}) with $\alpha=0.5$, $\beta=2$, and $s=0$. In addition to Bayes factors for each individual covariate, we also report the 95\% credible interval (CI). Although it might be tempting to use the CI to test hypotheses (e.g., rejecting the null hypothesis if the CI does not include the null value), in accordance with recent recommendations \citep{van2021cautionary,wagenmakers2020principle}, we use a more conservative approach, where the CI becomes relevant only after the \emph{BF} shows evidence for the alternative hypothesis.

\subsection*{Results}
We begin our reporting of the results by showing outcomes from objective performance metrics such as the performance of human and collaborative AI agents.
Additionally, we highlight behavioral measures related to the degree to which one agent interferes with the plan of the other agent. 
We then report the results from the subjective metrics based on the questionnaire responses.  
Finally, we examine human preferences for various types of collaborative agents and apply predictive models to determine which objective and subjective metrics best predict choice.

\subsubsection*{Objective Metrics}
\begin{figure}[h]
    \centering
    \includegraphics[width=1\textwidth]{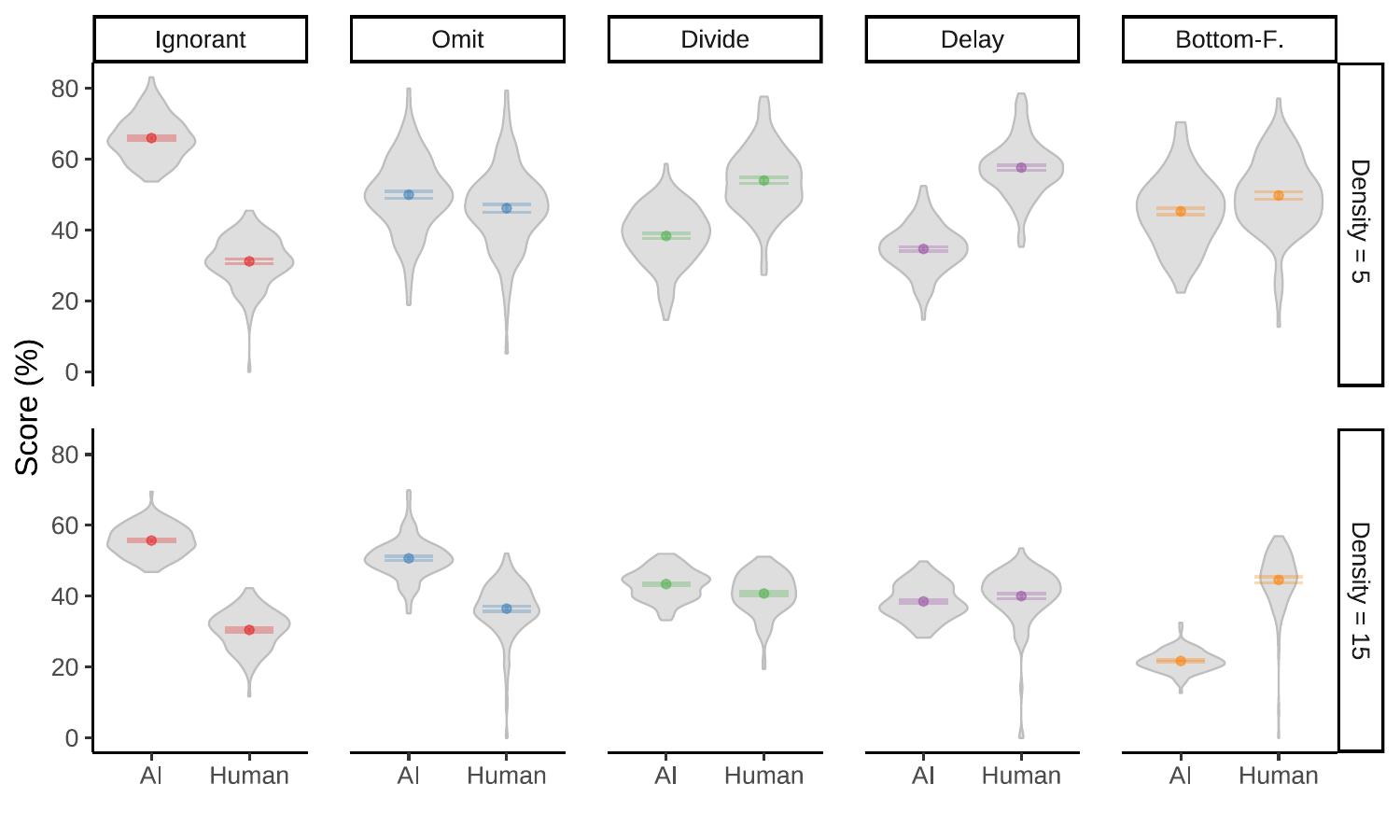}
    \caption{Performance of the human and AI player by AI agent type (columns) and target density (rows) in Experiment 1. Performance is assessed by a relative score: the total points scored relative to the total points that were available during game play. Gray areas visualize the distribution of proportional scores; error bars show the standard error of the mean.}
    \label{fig:propscores}
\end{figure}

\paragraph{Performance Differences}
Figure \ref{fig:propscores} shows the performances of the human player and the collaborative AI agent across different human-AI teams. The results show significant differences in individual human player and AI agent performance. The best-performing AI agent for both the low and high target density conditions was the Ignorant agent. At the same time, human performance was worst with the Ignorant agent. This shows that the Ignorant agent that ignored all human intentions and acted as a single player was effective in maximizing its own performance but had a negative impact on the performance of its human partner. 

In low target density (a maximum of 5 concurrent targets), participants achieved the highest and next highest performance when playing with the Delay and Divide agent, respectively. In fact, participants performed better with any of the experimental agents, relative to the Ignorant agent baseline, $BF_{10} > 100$. This shows that the agents that aimed to reduce conflict and performance differences best amplified human performance. In high target density (a maximum of 15 targets), human players performed best with the Bottom-Feeder agent. However, performance differences in the high target density condition were less pronounced, suggesting that the increased availability of targets to intercept led to human strategies that were less affected by the AI agent. 

\begin{figure}[h]
    \centering
    \includegraphics[width=1\textwidth]{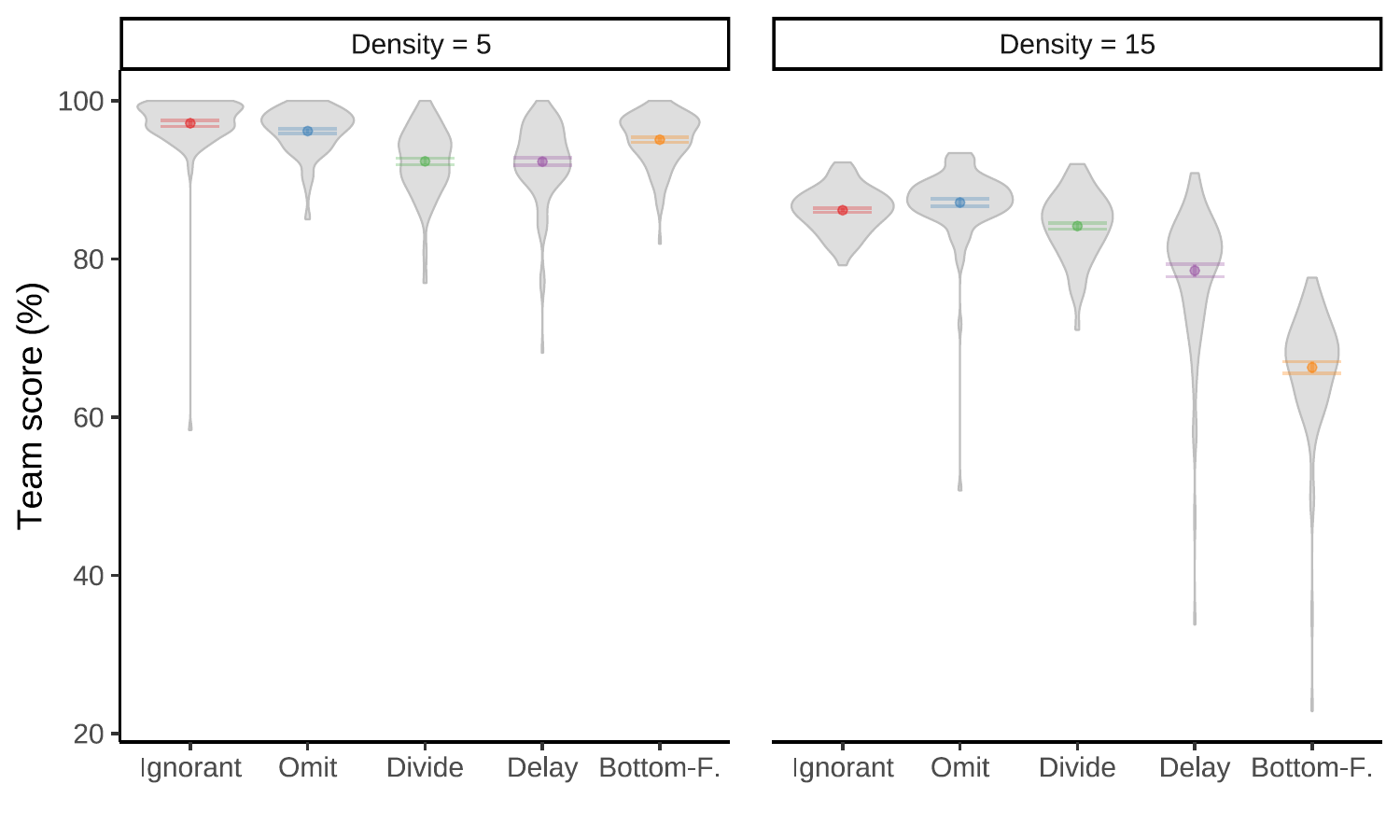}
    \caption{Mean team score by AI agent type and target density in Experiment 1. The team score is based on the sum of score of the AI agent and the human playing with that AI agent relative to the total value of points that was available during game play. Gray shading indicates the distribution of values, while error bars show the standard error from the mean.}
    \label{fig:teamperformancefig}
\end{figure}

Figure \ref{fig:teamperformancefig} shows the performance of the human-AI team where the score is combined across the human player and the AI agent. In the high target density condition, human+Omit teams slightly outperformed the human+Ignorant teams, although not statistically significant, $0.4 \leq BF_{10} < 1$. Thus, the additional design features of the Omit agent presumably caused human performance gains that were at least equally as high as the AI performance decreases. In the low target density condition, the best team performance was achieved with the Ignorant agent closely followed by the Omit agent. 

\paragraph{Other Behavioral Metrics}
We also examined several behavioral metrics that distinguish between the AI agents, focusing on measures related to conflict between the human player and the AI agent. One such metric, the number of AI 'steals', is defined as the number of times the AI agent intercepts a target initially pursued by the human player. Appendix Figure \ref{fig:steals} presents a visualization of these results. As expected, the Ignorant agent shows the highest number of interceptions of targets that the human intended to catch, since this agent disregards the human’s intentions and will pursue targets regardless of the human’s planned actions. 
Additionally, we analyzed the number of path intersections between the human player and the AI agent as another indicator of potential conflict. Path interceptions were operationalized as the presence of overlap in inter-agent movement trajectories since each agent has an avatar location in the game world and a location they are moving towards. A path intersection thus occurs when the agent's concurrent trajectories are intersecting.

\subsubsection*{Subjective Metrics: Questionnaire Responses}

\begin{figure}[h]
    \centering
    \includegraphics[width=1.0\textwidth]{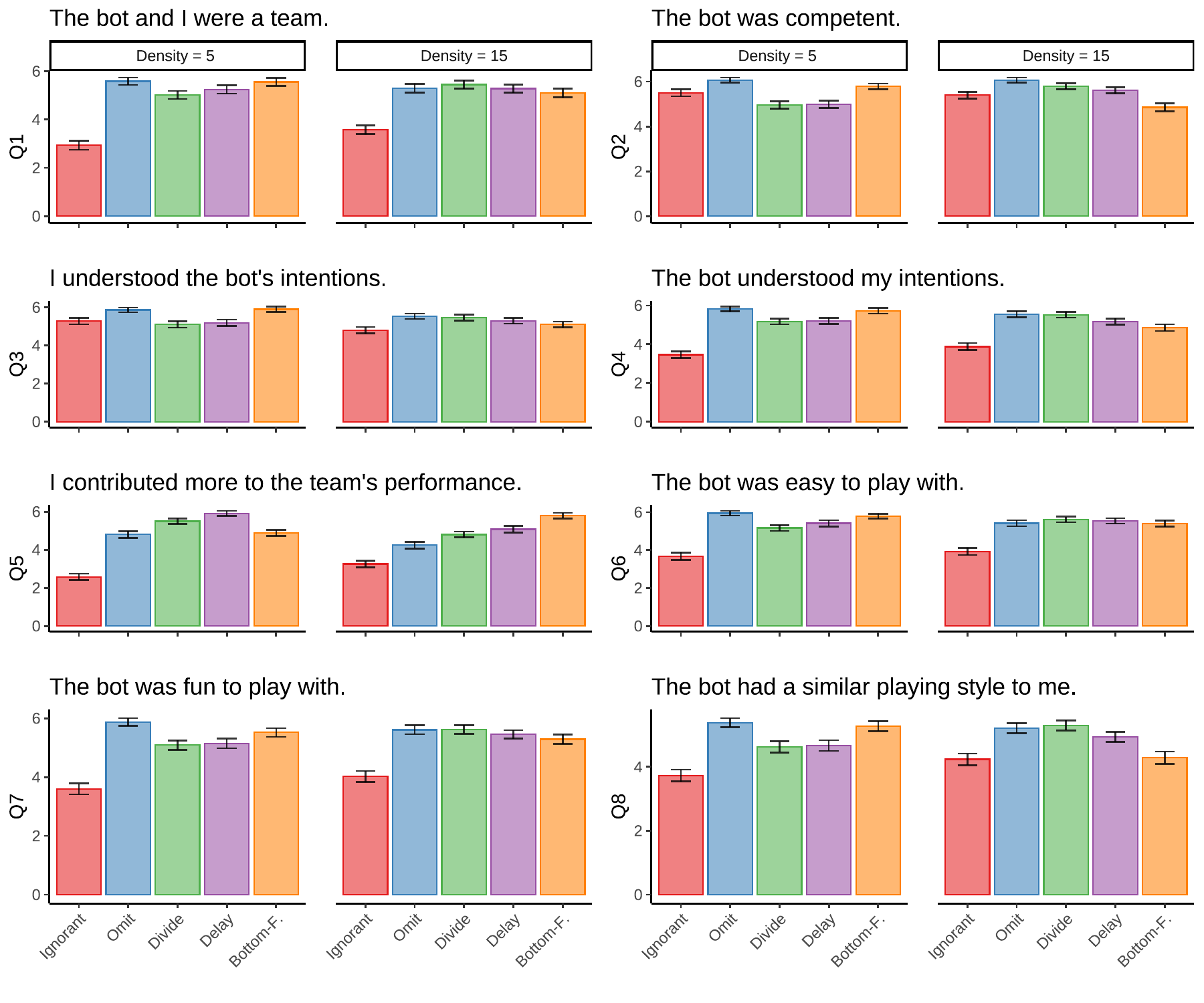}
    \caption{Mean questionnaire scores by AI agent type and target density in Experiment 1. Questions were rated on a 7-point Likert-scale. Error bars indicate the standard error from the mean.}
    \label{fig:questionnaireresponses}
\end{figure}

Figure \ref{fig:questionnaireresponses} shows the questionnaire results. The most general finding that holds for all items except Q3 (``I understood the bot's intentions'') is that there are significant differences in the ratings for the Ignorant agent, compared to all other agents. 
These statistical findings reveal a difference between the Ignorant agent and the other agents. However, for some items this pattern is more nuanced as demonstrated by the significant target density interaction effects observed for items Q2, Q5, Q8, $BF_{10} > 100$. 
Exceptions include Q2 (``The bot was competent'') where in the low-density condition the Ignorant, Divide, and Delay agents were rated equally well, while in the high-density condition the Bottom-feeder was rated worse than all other agents, including the Ignorant agent. 
Furthermore, comparisons in the high-density condition of Q8 (``The bot had a similar playing style to me'') show that the ratings of the Ignorant and Bottom-Feeder agents were roughly equally low, while all other agents received significantly higher ratings. Q3 evinced no differences in ratings across agents.  

Table \ref{tab:openendedresponses} shows example responses when participants were asked to explain their choice. Participants voice many of the human-centered design considerations in their open-ended responses. The theme of teaming was highly represented in our participants' open-ended responses. Participants frequently pointed out that the Ignorant agent was not being a good teammate. Appendix \ref{appendixopenended} provides a content analysis that confirms that the majority of open-ended responses focused on teaming. 

\begin{table*}[h] 
\footnotesize 
\centering
\caption{Examples of explanations provided in the open-ended surveys in Experiment 1. Participants typically referred to the bots using their color labels; for clarity, these references have been replaced with the corresponding bot names.}
\begin{tabular}{lp{4in}} 
\toprule
\textbf{Theme} & \textbf{Response} \\
\midrule \\
\vspace{0.12in}
Teaming & \begin{tabular}[c]{@{}p{4in}@{}}
``The [Omit] bot felt less like competition and more like a fellow teammate. When I would choose a target, even if it was originally planning on going to that target, it would get out of the way and let me grab the target. This seemed more aligned with two teammates working together than the other robot.'' 
\\ \\ 


``It felt like we were a team and just trying to collect as many circles as possible where as the [Ignorant] bot felt like it was competing against me and would go change direction based on the highest point circles rather than holding down an area of the platform like me and the [Bottom-Feeder] bot would.''
\end{tabular} \\
\midrule \\
\vspace{0.12in}
Likability & 
\begin{tabular}[c]{@{}p{4in}@{}}
``The [Divide] bot made playing the game easy and fun. The [Divide] bot spent most of its time in a quadrant away from where I was playing, allowing me not to feel crowded or pressured.'' \\
\end{tabular} \\
\midrule \\
\vspace{0.12in}
Intentionality & \begin{tabular}[c]{@{}p{4in}@{}}
``The [Bottom-Feeder] bot seemed to just let me get any of the targets that I wanted and didn't try to fight for them.''
\end{tabular} \\
\bottomrule
\end{tabular}
\label{tab:openendedresponses}
\end{table*}

\subsubsection*{Preferences for Collaborative Agents}
\begin{figure}[h]
    \centering
    \includegraphics[width=1\textwidth]{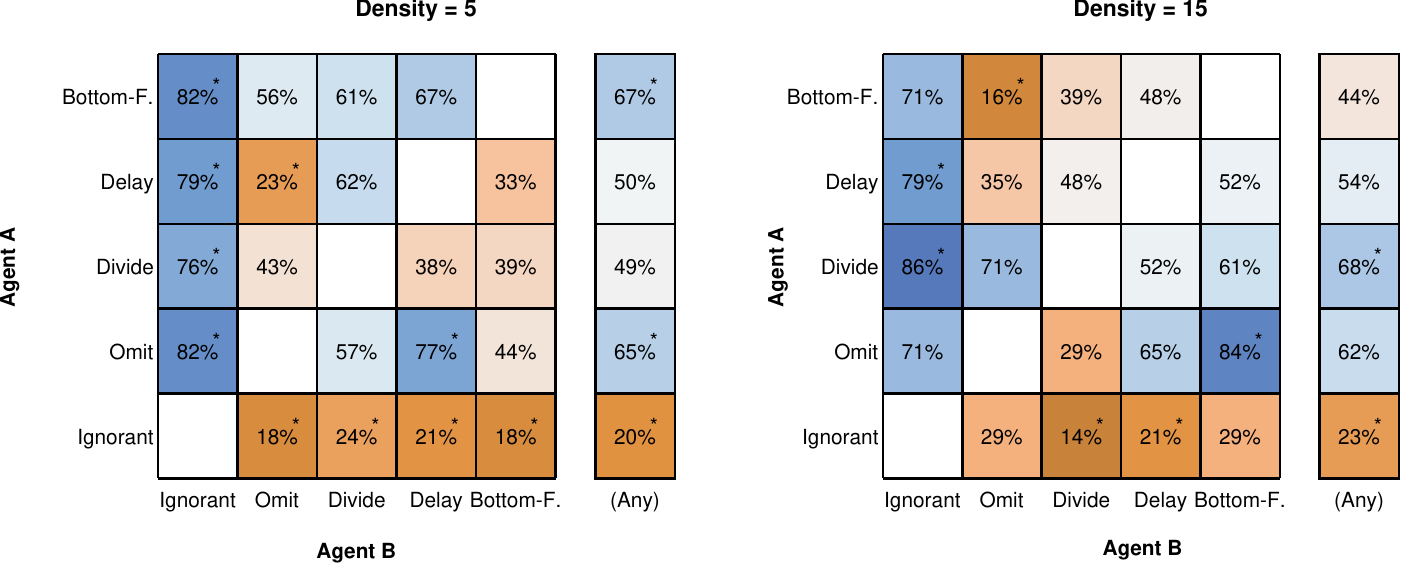}
    \caption{Choice preferences across pairs of agents for each target density condition in Experiment 1. Each matrix cell indicates the percentage of participants preferring the row-associated agent A over the column-associated agent B. For instance, 82\% of participants preferred the Omit agent over the Ignorant agent in the low target density condition. The side column shows the overall preference percentage for each row agent across all pairings. Asterisks denote choice percentages that significantly deviate from 50\%, indicated by a Bayes factor greater than 10. Note that results here are averaged across different presentation orders of the agents (e.g., agent A could have been presented first or second in the experiment).}
    \label{fig:choice_split}
\end{figure}

Figure \ref{fig:choice_split} shows participants' preferences for specific AI agents when they were paired with other agents. In the side columns, we observe that certain agents were consistently preferred regardless of the agent they were paired with. In the low target density condition, the most popular agent was the Bottom-Feeder, chosen 67\% of the time, followed closely by the Omit agent at 65\%. Conversely, the Ignorant agent was the least preferred, selected in only 20\% of pairings. For the high target density condition, the Divide agent became the most preferred, chosen 68\% of the time, with the Omit agent close behind at 62\%. Again, the Ignorant agent remained the least favored, chosen only 23\% of the time.

\subsubsection*{Predictive Models for Human Preferences}
What factors influence people's preferences for certain AI agents? To investigate this, we apply Bayesian logistic regression models to predict individual choices that people make in a pairwise comparison of two collaborative AI agents. The predictions are based on both objective metrics (e.g., performance-related metrics) and subjective metrics (e.g., Likert ratings that assess the subjective experience with the AI agents). Our approach is grounded in a Bradley-Terry framework \citep{bockenholt2001hierarchical, cattelan2012models}, where the likelihood of selecting Agent $X$ over Agent $Y$ in a pairwise comparison depends on the difference in their respective utility scores:
\begin{equation}
\log \left( \frac{P(\text{Choice} = X )}{P( \text{Choice} = Y )} \right) = U_x - U_y
\end{equation}
Here $U_x$ and $U_y$ represent the utility scores of agents X and Y, respectively. If these utilities are expressed as weighted sums of features, this model can be expressed by logistic regression:
\begin{equation}
\log \left( \frac{P(\text{Choice} = X | X,Y)}{P( \text{Choice} = Y | X,Y)} \right) = \beta_0 + \beta_1 (X_1-Y_1) + \beta_2 (X_2-Y_2) + \dots + \beta_n (X_n-Y_n)
\label{eq:choice}
\end{equation}
where \(X_i\) and \(Y_i\) represent the values of the \(i\)-th feature for Agents \(X\) and \(Y\), respectively, and \(n\) is the total number of features. This approach is centered on modeling each covariate in terms of the \emph{difference} between corresponding feature values of the two agents. The weights \(\beta\) indicate the relative influence of each feature, while \(\beta_0\) represents a bias term, accounting for any baseline preference for the first-presented agent in the pairwise comparison (i.e., assuming agents are presented in the order \(X\) followed by \(Y\)).

To estimate the weights in the logistic regression model of Equation \ref{eq:choice}, we use Bayesian methods. We separately apply the model to objective and subjective metrics, further breaking down the results by target density. For the objective metrics, we included features such as the human and AI scores, score inequality (defined as the absolute difference between human and AI scores), the number of AI steals, and the number of path intersections between the human and AI agent. For subjective metrics, we included Likert ratings from each of the 8 survey questions. Figure \ref{fig:regressionsplit} presents the posterior estimates for the \(\beta\) weights (see Appendix Table \ref{tab:posteriorSummariesOfCoefficients} for full results). A positive \(\beta\) weight indicates that a larger positive difference in feature values between Agents \(X\) and \(Y\) increases the likelihood of choosing Agent \(X\). 

\begin{figure}[h]
    \centering
    \includegraphics[width=0.8\textwidth]{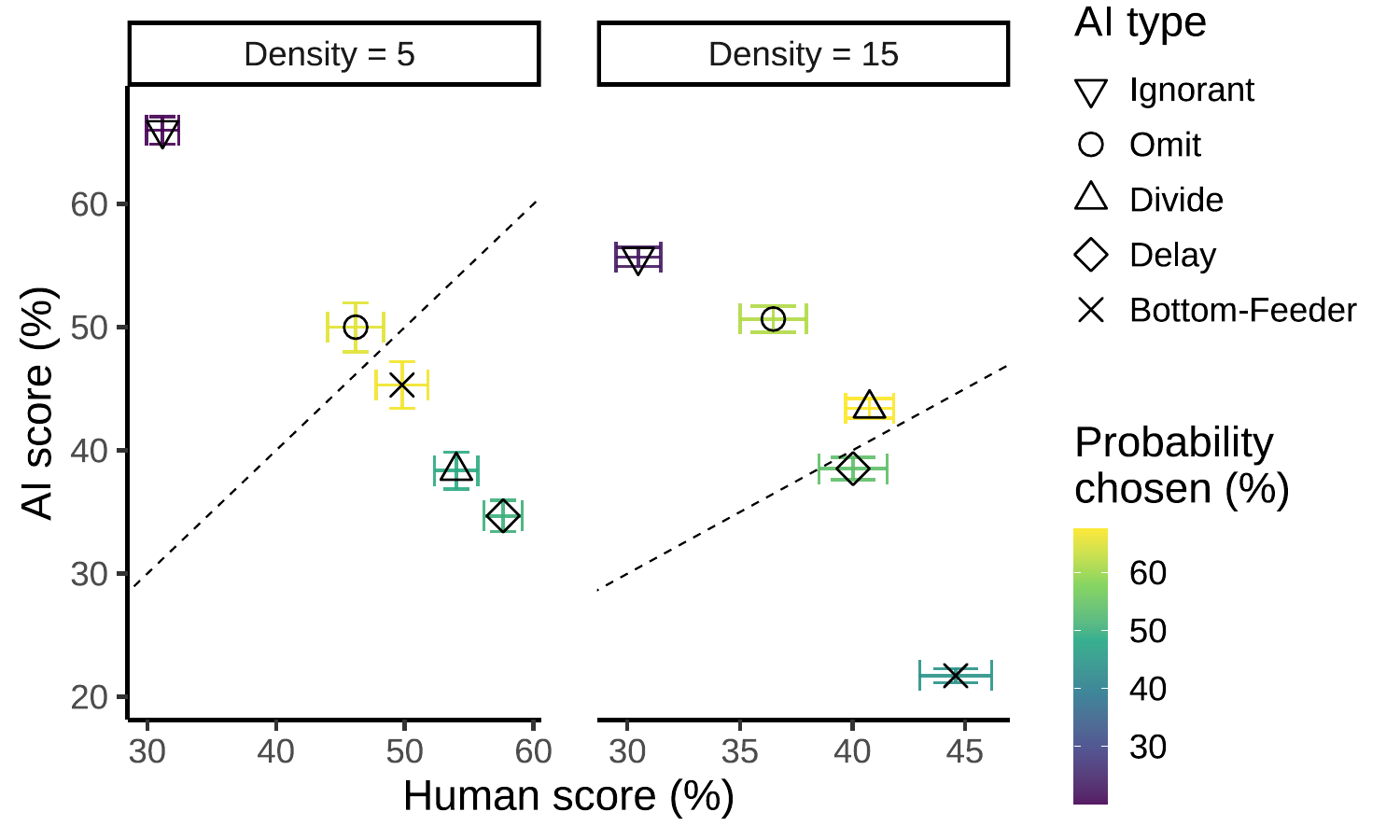}
    \caption{Scatter plot of mean relative AI agent score versus the human score in Experiment 1. The results are separated by type of human-AI team and  target density. Color is reflective of the probability of the AI type being preferred in pairwise match-ups. The diagonal represents the points of equal scores for the AI agent and the human player. The probability with which an agent is chosen appears inversely related to the distance from the diagonal, suggesting that human players have a preference for score equality. Error bars are indicative of the standard error from the mean.}
    \label{fig:inequalityandpreference}
\end{figure}

For the objective metrics, results indicate that there is no noteworthy effect of human and AI scores ($\emph{BF} < 3$) for either target density—human preferences are not driven by the performance of the AI agent or themselves. However, one key predictor is the score inequity ($\emph{BF} = 7$ and $\emph{BF} > 100$ for target densities 5 and 15, respectively). Agents that promote more equal performance between the human and the AI agent are preferred. To further illustrate this effect, Figure \ref{fig:inequalityandpreference} shows the effect of score inequality on human preferences. The results show that agents with human and AI scores that are more similar (i.e., closer to the diagonal lines representing equal scores) tend to be chosen more often. Additionally, there appears to be a bias towards human performance in the sense that the human outperforming the AI affects preferences less than the inverse.

Another key predictor is AI "steals", with agents scoring higher on these features being less preferred, although there is only convincing evidence for a non-zero estimate of this effect in the low target density model ($\emph{BF} > 100$), perhaps because in low density settings there is more opportunity for competitive interaction. 

For the subjective metrics, in the low target density condition, there is evidence for effects of predictors such as teaming (Q1), the AI’s ability to understand human intent (Q4), and a similar playing style (Q8) are influential ($\emph{BF} > 100$, $\emph{BF} = 5$, and $\emph{BF} > 100$, respectively). In high target density, predictors shift to understanding the AI's intent (Q3), the AI’s ability to understand human intent (Q4), ease of collaboration (Q6), enjoyment (Q7), and similarity in playing style (Q8) ($\emph{BF} = 5.2$, $\emph{BF} > 10$, $\emph{BF} > 100$, $\emph{BF} > 100$, and $\emph{BF} > 10$, respectively). 

\paragraph{Predictive Accuracy}
Another way to evaluate the model is through its ability to predict people's preferences. We applied a 10-fold cross-validation procedure where 90\% of the pairwise choice data was used to train the logistic regression and the remaining 10\% of the pairwise choice data was used to assess the accuracy of the model predictions. For the objective metrics, accuracy reached 62\% in both target density conditions. For subjective metrics, predictive accuracy was 84\% and 81\% for low and high target densities, respectively\footnote{When evaluating predictive accuracy using the AUC metric, the results are qualitatively similar. For the objective metrics, the AUC is 0.7 and 0.67 for the 5 and 15 target densities, respectively. For the subjective metrics, the AUC increased to .91 for both target densities}. These results suggest that subjective ratings are more predictive of choice and capture dimensions not represented in the objective metrics—at least within the set considered in this study.

\begin{figure}[h]
    \centering
    \includegraphics[width=1\textwidth]{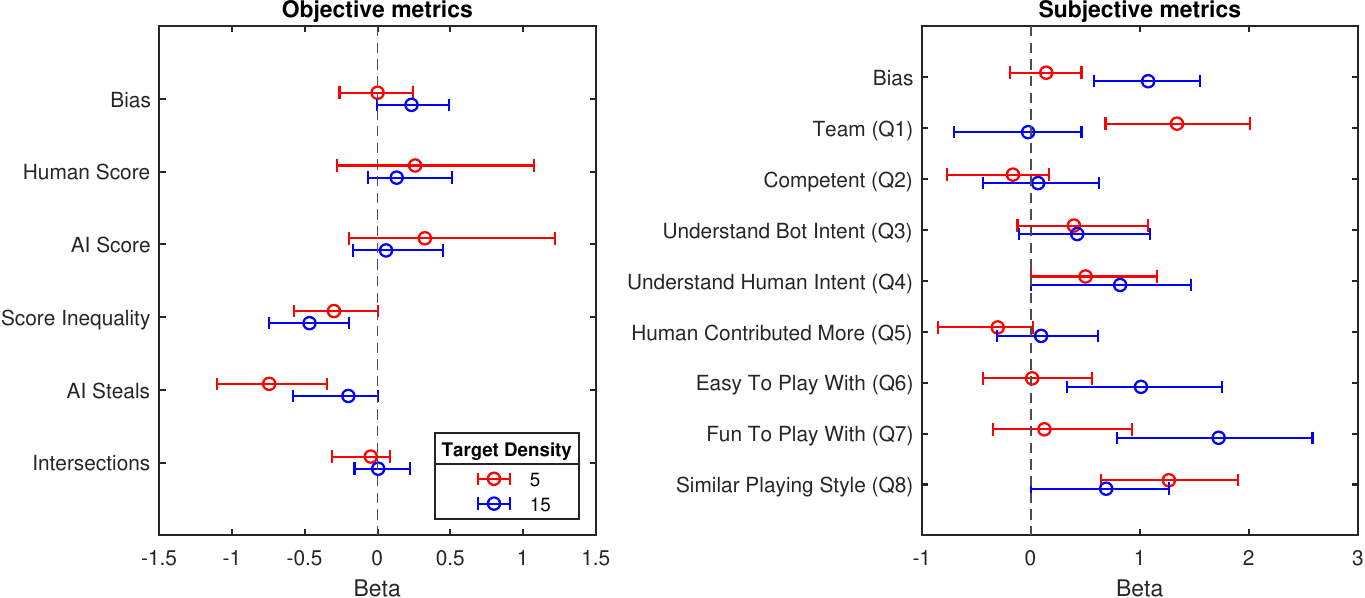}
    \caption{Posterior ($\beta$) coefficients of Bayesian logistic regression models predicting choice in Experiment 1. The coefficients are shown across objective and subjective metrics (left and right panels) and target densities (indicated by colors). Coefficient estimates can be thought of as weights for the importance of metrics in explaining choices. Error bars represent 95\% credible intervals.}
    \label{fig:regressionsplit}
\end{figure}

\begin{figure}[h]
    \centering
    \includegraphics[width=0.8\textwidth]{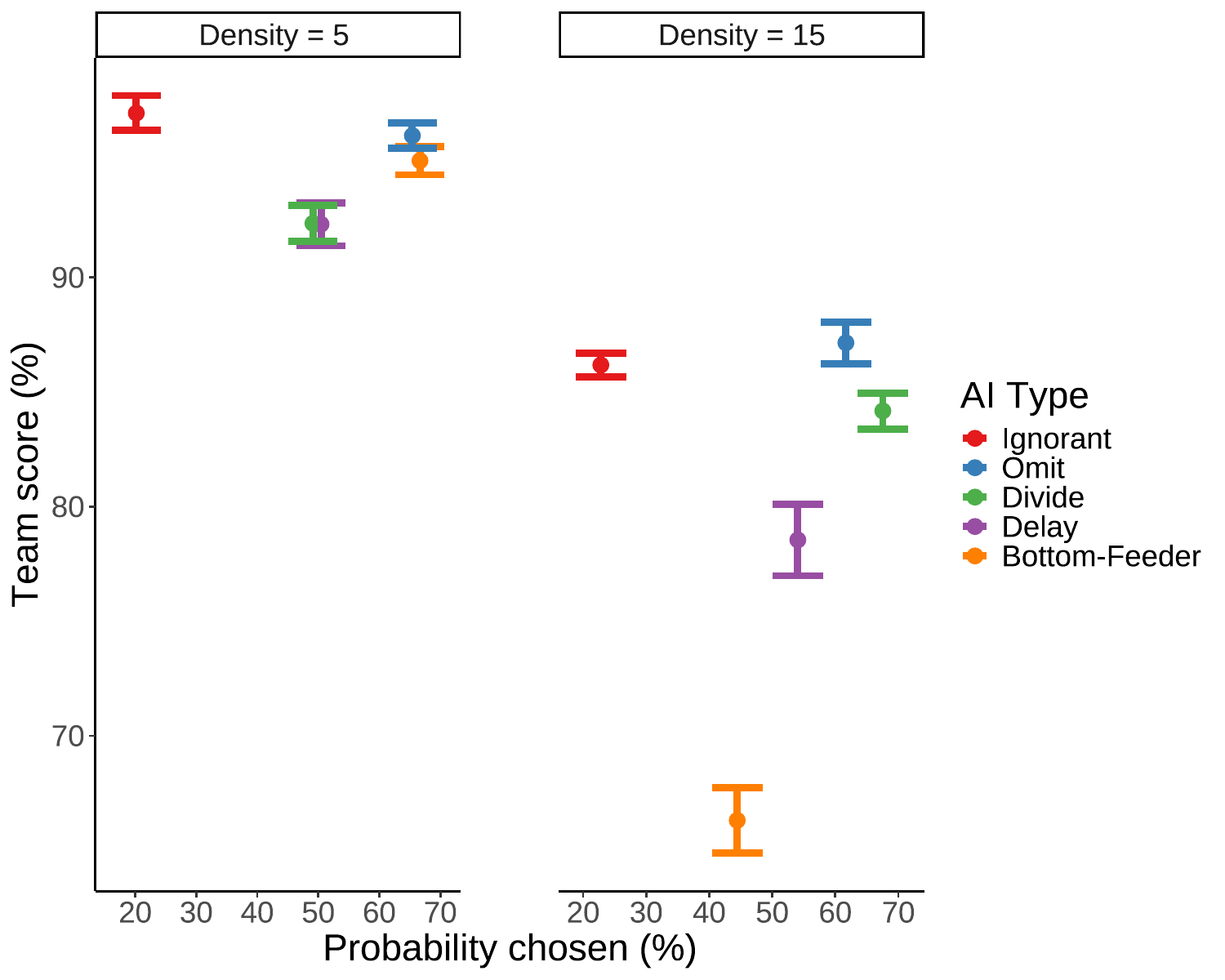}
    \caption{Scatter plot of team performance over probability of being chosen, by AI agent condition and target density conditions. Error bars reflect the standard error from the mean. Results are from Experiment 1.}
    \label{fig:perftradeoff}
\end{figure}

\paragraph{Trade-off Between Performance and Preference}
The results of the choice analysis show that human preferences for AI agents are driven by a number of factors other than the performance of the individual AI agents or human players. A visualization of this misalignment is shown in Figure \ref{fig:perftradeoff}. While certain agents, like the Ignorant agent, demonstrated high team scores by maximizing interception rates, this approach often led to lower human preference ratings due to competitive interactions that disregarded human intentions. Agents designed to reduce target competition, such as the Divide and Bottom-Feeder agents, were generally preferred by participants, especially under low target density conditions. The Divide agent’s area-based strategy minimized overlaps in target selection, enhancing collaborative ease, while the Bottom-Feeder agent focused on lower-value targets, allowing humans to prioritize high-value intercepts and feel a stronger sense of contribution.

Overall, these results show that selecting the best collaborative AI agent depends on the primary criteria for evaluation. If team performance is prioritized, agents like the Ignorant and Omit are reasonable choices. However, if human preference is the priority, agents such as Bottom-Feeder and Omit in low target density conditions, and Omit and Divide in high target density, would be preferred. From a multi-objective optimization perspective, the best collaborative AI agent balances these performance and preference goals.

\section*{Experiment 2}
The results of Experiment~1 indicated that participants’ choices of collaborative AI agents were influenced more strongly by subjective impressions than by either their own performance or the AI’s performance. One possibility is that certain aspects of the experimental design contributed to this outcome. In Experiment~1, after participants interacted with two agents, they were first presented with a questionnaire assessing the agents on multiple collaborative dimensions, and only afterward were they asked to choose their preferred agent. This ordering may have shaped participants’ subsequent choices by drawing attention to the specific traits emphasized in the questionnaire, thereby amplifying the influence of subjective factors relative to objective performance metrics. Additionally, because monetary compensation was fixed and independent of performance, participants may have been less motivated to prioritize performance-related considerations when selecting an agent.

Experiment~2 was designed to address these potential limitations by introducing two changes. First, we manipulated the order in which participants completed the preference choice and the questionnaire. This allowed us to test whether the act of rating an agent on specific traits before making a choice alters the relative weight of subjective and performance-based factors. Second, we introduced a performance-based incentive: participants received a \$2 bonus if their team’s cumulative score across the four game rounds ranked in the top 50\% of all participants. This bonus was intended to increase the salience of performance outcomes and test whether financial incentives would shift preferences toward higher-performing agents.

\subsection*{Methods}
The methodological framework for Experiment~2 closely mirrored that of Experiment~1, with two key modifications: (1) the introduction of a within-subject manipulation of the choice–questionnaire order, and (2) the addition of a performance-based bonus incentive.

\subsubsection*{Participants}
93 participants were recruited from the online participant recruitment platform Prolific. Ages ranged from 23 to 72 years (Mean = 43.36, SD = 11.72), with 42\% identifying as female, 57\% as male, and 1\% choosing not to disclose gender. All participants resided in the United States and had not participated in Experiment 1. The study was conducted remotely on participants’ personal computers. Each participant received a base payment of 5 USD for completing the approximately 25-minute study, corresponding to an average base rate of 12.43 USD/hour. In addition, participants were eligible for a 2 USD bonus if their cumulative team score across the four main rounds of gameplay ranked in the top 50\% of all participants within their assigned conditions. In total, 40 participants received bonus payments, bringing the final average hourly compensation rate to 14.38 USD/hour.

\subsubsection*{Procedure and Design}
The game environment and agents were identical to Experiment 1. The study followed the same experimental manipulations and procedures as in Experiment 1. 

The key change was the order manipulation, implemented within participants across the two blocks: each participant experienced both the Survey→Choice (after one block) and the Choice→Survey order (in the other block). Participants were randomly assigned to one of two conditions. In condition 1, participants completed Survey→Choice after Block 1 and Choice→Survey after Block 2 (N = 48), whereas participants in condition 2 completed Choice→Survey after Block 1 and Survey→Choice after Block 2 (N = 45). Thus, order varied within participants (by block), and the between-participant factor was which order occurred first. All other procedures—including instructions, tutorial and comprehension checks, rating scales, free-text rationales, and end-of-study feedback—were unchanged from Experiment 1.

\subsection*{Results}
The analyses of Experiment~2 examined whether altering the order of the choice and questionnaire tasks influenced the relative weight of performance-based (objective) versus perception-based (subjective) factors in participants’ preferences, and whether the introduction of a performance-based bonus shifted these preferences. For additional experimental results, see Appendix \ref{appendixmoreresultsexp2}.

\subsubsection*{Predictive Models for Human Preferences}
We applied the same Bayesian logistic regression framework used in Experiment~1 to model choice behavior. Models were fit separately for each order—Survey→Choice and Choice→Survey—with objective and subjective predictors analyzed independently.

\begin{figure}[h]
    \centering
    \includegraphics[width=1\textwidth]{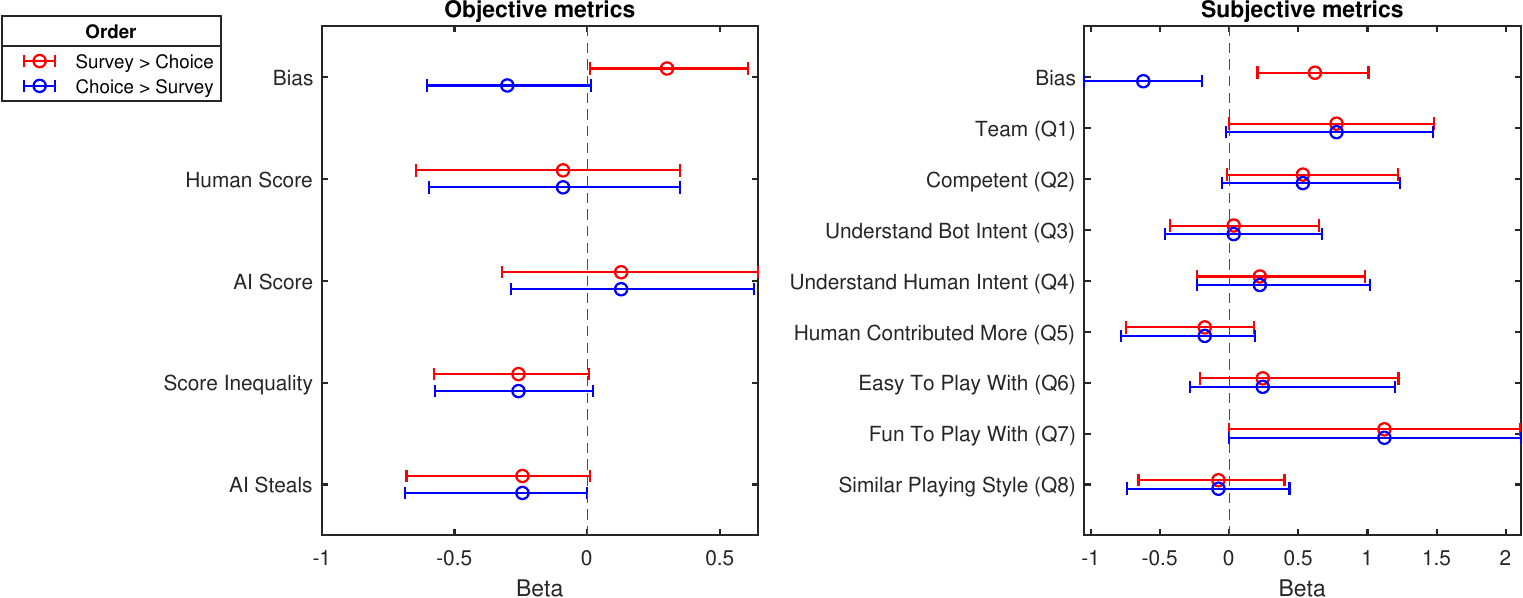}
    \caption{Posterior ($\beta$) coefficients of Bayesian logistic regression models predicting choice. The coefficients are shown across objective and subjective metrics (left and right panels) and survey order (indicated by colors). Coefficient estimates can be thought of as weights for the importance of metrics in explaining choices. Error bars represent 95\% credible intervals.}
    \label{fig:regressionsplitexp2}
\end{figure}

As shown in Figure \ref{fig:regressionsplitexp2}, there were no meaningful differences between the two order conditions in the predictive importance of either objective or subjective measures. In other words, the ordering of the survey and forced-choice tasks did not substantially change the relative influence of performance-based or subjective factors on agent preference.

One notable difference between the conditions emerged in the bias term. When the survey preceded the choice, participants were more likely to select the first agent presented (consistent with Experiment~1). When the choice preceded the survey, however, participants tended to select the second (most recently presented) agent. This pattern is consistent with a primacy/recency effect \citep{mantonakis2009order}: if the choice immediately follows the second agent’s presentation, recency may drive preference, whereas an intervening activity—such as answering survey questions—appears to shift attention back toward the first-presented agent.   

\subsubsection*{Predictive Accuracy}
For the survey→choice condition, the model achieved predictive accuracies of 62\% for objective metrics and 78\% for subjective metrics—closely matching the performance levels observed in Experiment~1 (62\% and 84\%, respectively). When the order was reversed (choice→survey), predictive accuracies were nearly identical: 62\% for objective metrics and 79\% for subjective metrics.

\subsubsection*{Preferences for Collaborative Agents}
To evaluate whether the introduction of performance-based incentives in Experiment~2 shifted participants' preferences for collaborative agents, we compared marginal choice percentages with those observed in Experiment~1. In Experiment~2, agents were chosen in the following order from most to least preferred: Omit (74\%), Delay (43\%), Divide (42\%), Bottom-Feeder (59\%), and Ignorant (31\%). In Experiment~1, the order was Omit (64\%), Divide (59\%), Delay (58\%), Bottom-Feeder (49\%), and Ignorant (22\%). The overall ranking of agents was similar across experiments, with the only change being a swap in relative positions of the Delay and Divide agents, whose marginal preference rates were closely matched in both studies.

\subsection*{Discussion}
The findings from Experiment 2 indicate that the ordering of choice and survey responses had negligible influence on the predictive power of either objective or subjective measures. The stability of these accuracies across conditions suggests that participants’ evaluation strategies and choice behaviors were not meaningfully altered by whether they rated the agents before or after making their selection. In both orderings, subjective ratings remained substantially more predictive of choice than objective performance metrics, consistent with the patterns observed in Experiment 1.

In addition, the absence of an incentive effect in this study suggests that, under the modest stakes offered here, participants’ choices of collaborative agents were not influenced by the prospect of a performance-based bonus. Even when a financial reward was contingent on achieving higher team scores, participants continued to prioritize agents they perceived as supportive, complementary, and considerate of their actions over those that simply maximized team performance.  However, it remains possible that larger or more salient incentives could shift these priorities, leading participants to place greater weight on performance outcomes when selecting collaborators. Future work should focus on high-stakes settings to determine whether and how stronger incentives alter human-AI preference trade-offs.

\section*{General Discussion}
In this study, we sought to understand the traits that enable AI collaborators to be likable and competent partners in human-in-the-loop multi-agent task settings. To study this, we designed a behavioral experiment where humans play alongside an AI collaborator in a novel decision-making task. The most preferred agents were those that performed well on subjective evaluations of their collaborative abilities. In contrast to this, productivity on objective metrics did not prove to be a strong predictor of human preferences for collaborative AI agents.
Our study makes the following key contributions to human-AI collaboration research:
\begin{enumerate}
    \item \textit{People prefer AI collaborators that enable meaningful human contribution}. In our study, collaborative agents that showed greater performance differences compared to participants were generally less preferred. This observation aligns with previous research, which demonstrated that inequity aversion can enable groups of autonomous agents to maintain cooperative behavior \citep{hughes2018inequity}. However, our study is the first to demonstrate this effect in a human-AI collaboration context. We also found that people are more averse to falling behind the AI than they are to getting ahead of the AI. This tendency highlights some asymmetry that could reflect a human preference to lead the human-AI team. 
   \item \textit{People prefer AI collaborators that are considerate of human intentions}. Participants showed a clear aversion to perceived intrusions into the tasks they delegated for themselves. Likewise, participants' preferences were not significantly influenced by the AI's actual performance or their subjective assessment of its performance. Our findings also demonstrate how a considerate, or human-aware, AI collaborator does not always take away from the objective performance of the human-AI team. We even found that considerate AI collaborators contribute to collective performance as well as egocentric AI collaborators. For example, our results showed that the human-Omit team performed as well as, and sometimes better than, the team that paired a human with a performance-maximizing agent. This evidence of complementarity in human-AI collaboration—where human-aware algorithms boost individual and team performance—suggests that more sophisticated human-aware AI systems may exceed the performance of teams paired with purely performance-optimized agents. Taken together, our findings motivate reasons to design collaborative AI that better supports human capabilities, as this may increase user adoption and human-AI team performance. 
   
   \item \textit{The best collaborative agents are well adapted to their environment}. We found that participants generally favored simpler agents (Omit, Bottom-Feeder)\footnote{These agents are nearly identical in conditions of scarce resources. Both agents avoid intercepting targets for which the player has shown an intention to intercept. The agents only differ in their prioritization of targets of different values: Omit prioritizes high-value targets, while the Bottom-Feeder prioritizes the lowest-value objects.} in resource-constrained environments (low target density). On the other hand, people preferred more complex agents (Divide) in resource-rich (high target density) environments. Based on participant ratings and analysis of open-ended feedback, we interpret these preference shifts as an outcome of these AI agent policies being differentially more effective for some task environments over others. For instance, under resource-constrained settings, the Bottom-Feeder allows the human to pursue all high-value targets while focusing on the objects the human is unlikely to prioritize. This strategy becomes disadvantageous in resource-rich settings, as it now becomes apparent that low-value targets are best ignored under these conditions. Our results demonstrate how people are sensitive to agent strategies that better complement the available actions and rewards in a given task environment. 
      
   \item \textit{Small changes in the algorithm can potentially improve AI's collaborative abilities}. The different collaborative AIs in this study can be thought of as simple modifications to the inputs of an existing system. For example, we modified what our base agent can perceive about the world so that a Divide-and-Conquer strategy emerges in its behavior. Input modifications of the type we present in this work provide a possibility to enhance existing systems without requiring significant changes to their underlying algorithms. As such, our approach could improve new and existing collaborative systems by transforming how AI collaborators perceive and interact with their environment.  

   \item \textit{People’s preferences for AI collaborators remain stable even when modest performance incentives are introduced}. In Experiment~2, introducing a performance-based bonus and varying the order of preference and survey tasks did not meaningfully shift the balance between subjective impressions and performance metrics in predicting choice. Participants continued to favor agents perceived as supportive and considerate of their actions over those that maximized team scores. These findings suggest that, under low-stakes conditions, human-centered design features may outweigh performance considerations in shaping AI collaborator preferences.
\end{enumerate}

\subsection*{Limitations and Future Work}
Future research could explore reinforcement learning agents as collaborators. Due to the nature of our behavioral experiments, in which the agents and the entire game engine were rendered client-side (i.e., in the participant's browser), implementing MARL algorithms to act in real-time with human participants is computationally challenging. As a result, we employed a utility-maximizing interception algorithm that included several heuristic modifications to ensure the possibility of real-time collaborative behavior. 

In addition, more work is needed to translate and validate qualitative design principles that enhance collaboration into solutions that can either be integrated with existing algorithms or give rise to new, human-centered algorithms. Our research attempted to achieve this by constructing and evaluating algorithmic manipulations that emulate some of these theoretical insights. Further integration of human-centered design insights and machine learning research may require one of two approaches. In the short term, it may require more research that maps human subjective preferences to observable AI behaviors. This can look like evaluating whether certain behaviors consistently map to specific perceived traits. In the long term, a computational account of human preference formation in human-AI collaborative settings may be required to anticipate human preferences in new task settings.

\section*{Conclusion}\label{sec13}
 According to our analysis, people prefer to collaborate with algorithms that enable them to contribute meaningfully to the team, with an AI teammate that complements their actions rather than dominating the interaction. We found evidence of preferences being informed by a tendency to favor inequity aversion, implying that the collaborative AI's capabilities should be comparable to those of the human user. We also found strong support for preferences in agents that show a tendency to defer to the human. One avenue for achieving such approximations of collaborative abilities is through labor divisions that emphasize user autonomy, for example, through task delegation or spatial separation. Our experimental paradigm can be leveraged to iteratively improve the collaborative abilities of new and existing systems alike.

\backmatter

\section*{Declarations}

\paragraph{Funding} This research was funded by the Honda Research Institute (HRI) with the funding awarded to Steyvers. HRI was involved in the design of the study and the analysis of the data. 

\paragraph{Competing interests} The authors declare that they have no competing interests

\paragraph{Ethics approval and consent to participate} 
The study protocols were approved by the Institutional Review Board of the University of California, Irvine (IRB \#4527). Informed consent was obtained from each participant before the study commenced.

\paragraph{Availability of data and materials}
The behavioral datasets generated during the current study are available at OSF, \url{https://osf.io/ybweq/?view_only=cb4d4c7ac0b848b79b6ae8c7b09278cc}.

The experiment code is available on GitHub, \url{https://github.com/MADLABatUCI/target-intercept-collab-ai}.

\paragraph{Open Practices Statement} 
The behavioral datasets generated during the current study are available at OSF, \url{https://osf.io/ybweq/?view_only=cb4d4c7ac0b848b79b6ae8c7b09278cc}.

The experiment code is available on GitHub, \url{https://github.com/MADLABatUCI/target-intercept-collab-ai}. Navigate to the README file to play with each of the respective agents.

The experiments reported in this article were not pre-registered.

\paragraph{Consent for publication}
Not applicable

\paragraph{Author contribution} LM, SK, JA, MS, YY, DT, and MS designed the study. SK developed and ran the behavioral experiments. LM, SK, JA, and MS analyzed the data. LM, SK, and MS were major contributors in writing the manuscript. All authors read and approved the final manuscript.

\paragraph{Acknowledgments}
Not applicable

\noindent

\newpage

\newpage

\begin{appendices}
\counterwithin{figure}{section}  
\renewcommand{\thefigure}{B\arabic{figure}}  
\setcounter{figure}{0}  



\section{Questionnaire}

\begin{table*}[h] 
\centering
\caption{Survey Questions}
\label{tab:survey-summary}
\begin{tabular}{lp{1.0\textwidth}} 
\midrule
Q1 & ``The bot and I were a team.''\\
Q2 & ``The bot was competent.''\\
Q3 & ``I understood the bot’s intentions.''\\ 
Q4 & ``The bot understood my intentions.''\\
Q5 & ``I contributed more to the team’s performance.''\\
Q6 & ``The bot was easy to play with''\\
Q7 & ``The bot was fun to play with.''\\ 
Q8 & ``The bot and I had a similar playing style.''\\
\bottomrule
\end{tabular}
\end{table*}

\section{Search Algorithm}
\label{searchalgorithm}
The search algorithm computes all possible interception sequences involving up to three targets, updating the positions of both the AI player and the targets throughout the sequence. For each sequence, a total value is calculated by adding the intercepted objects' points. The interception plan selected is based on the sequence with the highest cumulative point gain. 

The planning algorithm incorporates a heuristic to account for the possibility of future objects entering the scene. More specifically, when planning for the first, second, and third interception, the algorithm discounts the value of future interceptions by a factor of $\alpha^K$ where $K$ is the estimated number of new objects that will enter the scene by the time a new interception is planned and $\alpha$ is a discounting parameter. Simulations determined that $\alpha=0.9$ led to good performance levels across different target densities. The search algorithm recomputes the plan in real time, allowing for changes in the game state, such as when a new target enters the game view or when a human player intercepts a target (preventing the AI player from intercepting it). 

The algorithm includes a stability parameter to prevent the AI player from making erratic moves in response to new incoming targets. The stability causes the AI player to only change its current plan if a new plan is at least 20\% better in expected point value than the current one. This threshold was determined through pilot studies, for which we found that lower thresholds resulted in participants perceiving the AI as overly erratic. Conversely, higher thresholds led the AI to appear overly rigid in its decision-making. By incorporating this stability component, the AI's behavior becomes more predictable, allowing human players to plan around the AI player.

\section{Additional Results Experiment 1}
\label{appendixmoreresults}
\renewcommand{\thefigure}{C\arabic{figure}}  
\setcounter{figure}{0}  

\begin{figure}[H]
    \centering
    \includegraphics[width=1\textwidth]{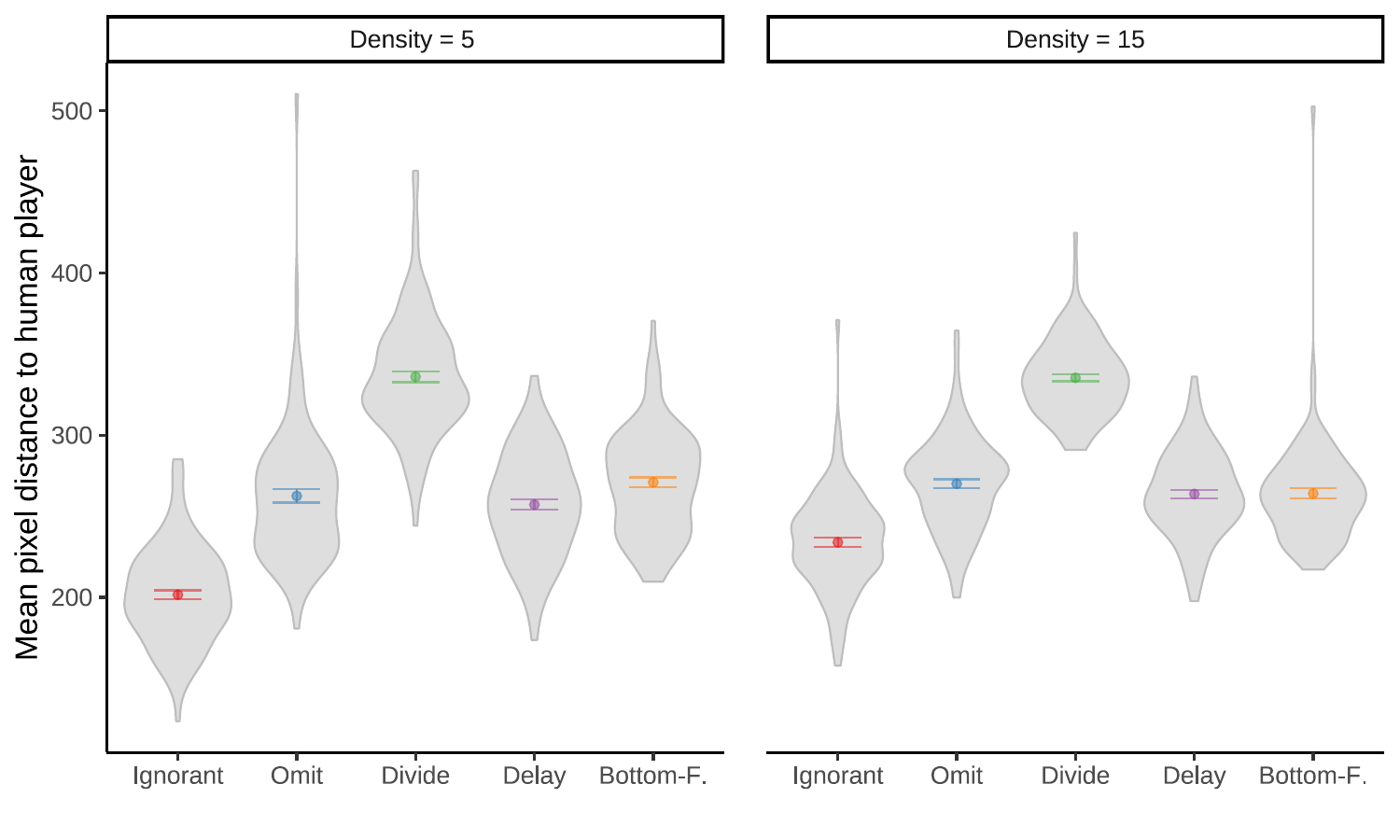}
    \caption{Mean pixel distance between the human and AI player separated by AI agent type and target density. Gray shading visualizes the distribution of individual mean pixel distances, while error bars show the standard error of the mean.}
\end{figure}

\begin{figure}[H]
    \centering
    \includegraphics[width=1\textwidth]{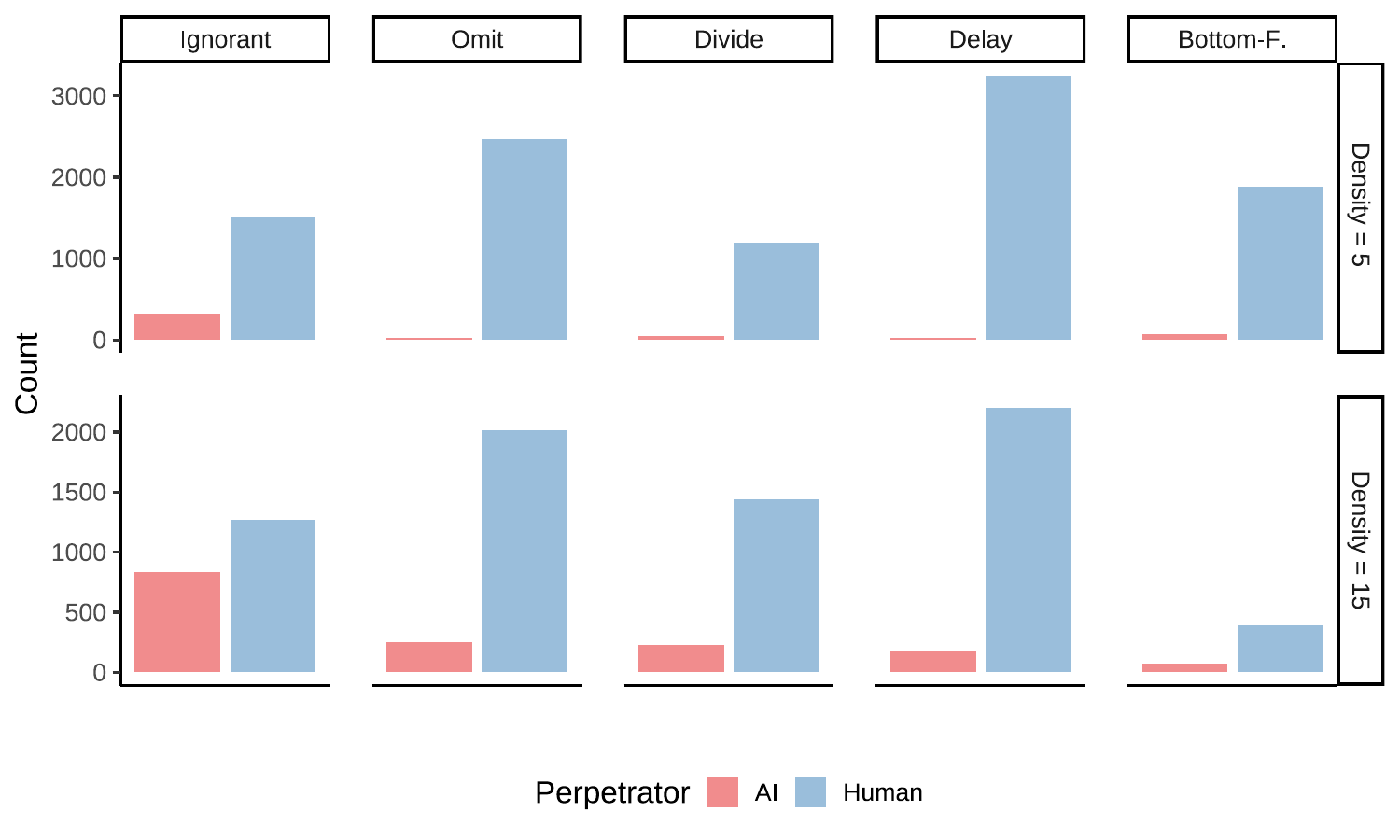}
    \caption{Number of stealing occurrences for each player split by AI agent type and target density. A steal is defined as an instance where one agent marked a target before the other agent with the latter agent intercepting that target.}
    \label{fig:steals}
\end{figure}

\section{Analysis of Open-Ended Responses Experiment 1}
\label{appendixopenended}
At the end of each experimental block, participants provided free-response feedback to identify aspects they liked or disliked about the AI agent's performance. 
The goal of the analysis in this section is to understand if participants' open-ended responses share characteristics that are present in the statements featured in the Likert questionnaire.

We first created single-word labels of participants' descriptions using an approach that combined human raters and  natural language processing (NLP).
This analysis began with four raters classifying the comments as either positive or negative. Our four human raters then described the agents using a single word or short phrase. We then utilized an NLP procedure to ensure that words that appeared repeatedly were standardized to a single term, and multi-word aspects were condensed into concise, single-word labels. We then applied human validation to ensure that these single-word labels were properly balanced into a positive or negative classification.

Subsequently, three raters categorized these extracted terms according to the eight items on the evaluation scale, allowing us to compare participants' subjective feedback with their quantitative ratings from Table \ref{tab:survey-summary}. The inter-rater reliability was assessed using the intraclass correlation coefficient, which yielded a high score of 0.868, indicating strong agreement among raters.

The results shown in Figure \ref{fig:open-ended-results} highlight that participants' open-ended sentiments describe the Ignorant agent as a poor teammate. Specifically, the Ignorant agent is represented negatively on teaming phrasing related to Q1 of the questionnaire. The ``teaming'' construct is frequently mentioned across descriptions of all the agents. On the other hand, the open-ended responses infrequently mentioned terms related to Q6, Q7, and Q8 which were predictive features in the regression model for the choice data (see Figure \ref{fig:regressionsplit} in the main paper). These results suggests that only a few salient dimensions such as teaming might be reported in the open-ended responses.

\renewcommand{\thefigure}{D\arabic{figure}}  
\setcounter{figure}{0}  
\begin{figure}[H]
    \centering
    \includegraphics[width=0.7\textwidth]{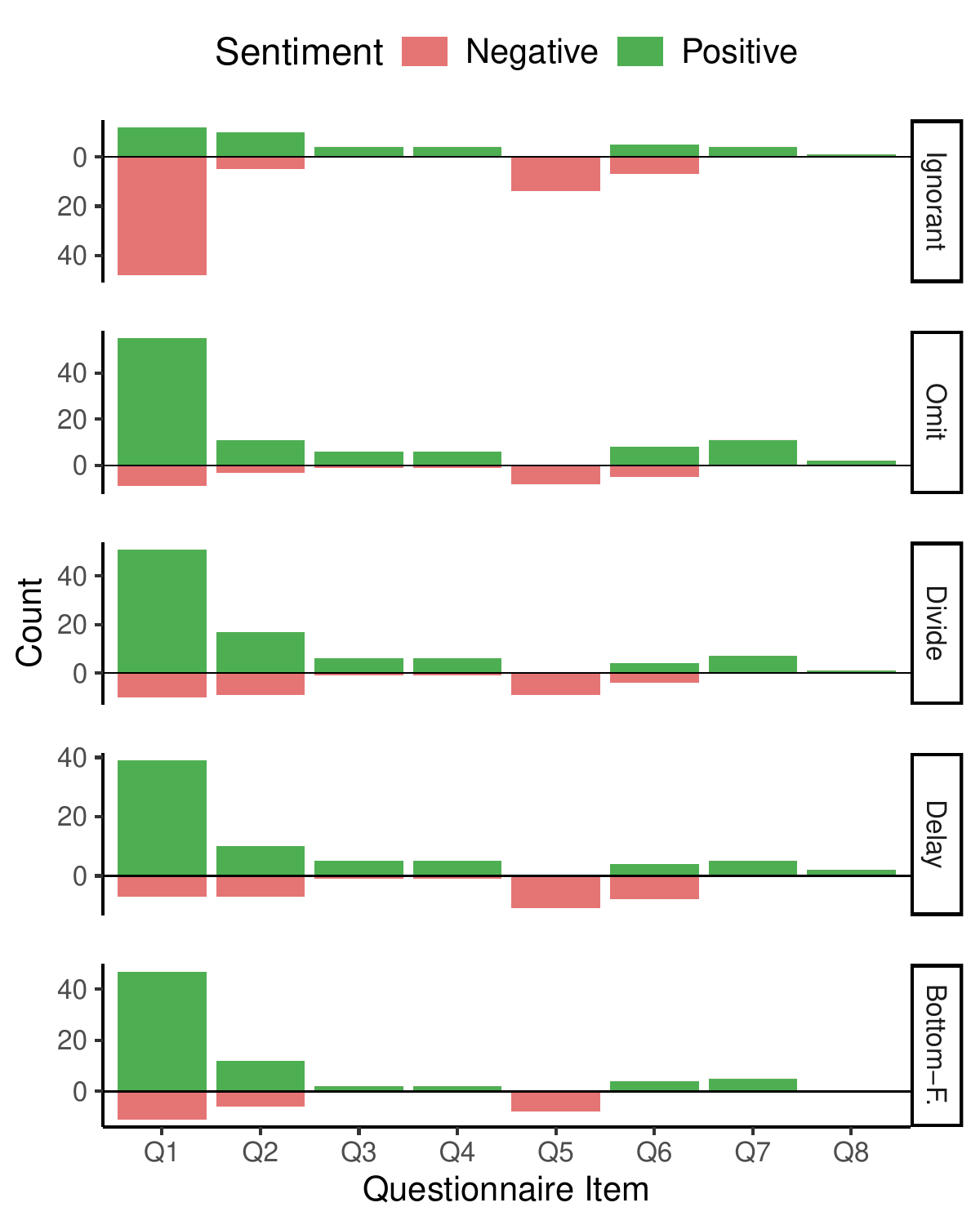}
    \caption{Sentiments for each agent based on the open-ended response section in relation to themes from the questionnaire items. Each statement contributed a single count to either positive or negative sentiments. Each open-ended statement was systematically coded as either being positive (green bars) or negative (red bars).}
    \label{fig:open-ended-results}
\end{figure}

\FloatBarrier
\newpage
\counterwithin{table}{section}   
\renewcommand{\thetable}{E\arabic{table}}
\setcounter{table}{0}
\section{Regression Modeling Results}

\begin{table}[htb]
\centering
\caption{Posterior Summaries of Coefficients Predicting Choice in Experiment 1}
\label{tab:posteriorSummariesOfCoefficients}
{
\begin{tabular}{lrrrrr}
\toprule
\multicolumn{1}{c}{} & \multicolumn{1}{c}{} & \multicolumn{1}{c}{} & \multicolumn{1}{c}{} & \multicolumn{2}{c}{95\% Credible Interval} \\
\cmidrule(lr){5-6} 
Model \& Coefficient & BF$_{inclusion}$ & Mean & SD & Lower & Upper \\
\cmidrule[0.4pt]{1-6}
Target Density = 5, Objective Metrics\\ 
\hspace{15pt}Bias & $1.000$ & $-1.386\times10^{-4}$ & $0.131$ & $-0.262$ & $0.246$  \\ 
\hspace{15pt}Human Score & $1.213$ & $0.258$ & $0.397$ & $-0.281$ & $1.076$  \\ 
\hspace{15pt}AI Score & $1.433$ & $0.325$ & $0.431$ & $-0.196$ & $1.221$  \\ 
\hspace{15pt}Score Inequality & $7.108$ & $-0.300$ & $0.173$ & $-0.577$ & $0.000$  \\
\hspace{15pt}AI Steals & $8101.536$ & $-0.745$ & $0.190$ & $-1.105$ & $-0.349$  \\ 
\hspace{15pt}Intersections & $0.978$ & $-0.047$ & $0.101$ & $-0.315$ & $0.083$  \\
\\
Target Density = 5, Subjective Metrics\\ 
\hspace{15pt}Bias & $1.000$ & $0.140$ & $0.177$ & $-0.193$ & $0.463$  \\ 
\hspace{15pt}Team (Q1) & $7407.462$ & $1.338$ & $0.359$ & $0.683$ & $2.009$  \\ 
\hspace{15pt}Competence (Q2) & $1.372$ & $-0.165$ & $0.262$ & $-0.769$ & $0.167$  \\ 
\hspace{15pt}Understand Bot Intent (Q3) & $2.949$ & $0.393$ & $0.369$ & $-0.124$ & $1.076$  \\ 
\hspace{15pt}Understand Human Intent (Q4) & $5.006$ & $0.500$ & $0.369$ & $-0.003$ & $1.159$  \\ 
\hspace{15pt}Human Contributed More (Q5) & $2.749$ & $-0.305$ & $0.285$ & $-0.855$ & $0.020$  \\ 
\hspace{15pt}Easy To Play With (Q6) & $0.917$ & $0.011$ & $0.224$ & $-0.436$ & $0.560$  \\ 
\hspace{15pt}Fun To Play With (Q7) & $1.087$ & $0.123$ & $0.320$ & $-0.344$ & $0.922$  \\ 
\hspace{15pt}Similar Playing Style (Q8) & $5291.962$ & $1.264$ & $0.337$ & $0.642$ & $1.901$  \\ 
\\
Target Density = 15, Objective Metrics\\ 
\hspace{15pt}Bias & $1.000$ & $0.233$ & $0.128$ & $-0.007$ & $0.491$  \\ 
\hspace{15pt}Human Score & $1.329$ & $0.131$ & $0.174$ & $-0.069$ & $0.512$  \\ 
\hspace{15pt}AI Score & $0.768$ & $0.058$ & $0.143$ & $-0.172$ & $0.448$  \\ 
\hspace{15pt}Score Inequality & $272.313$ & $-0.468$ & $0.146$ & $-0.748$ & $-0.199$  \\
\hspace{15pt}AI Steals & $2.454$ & $-0.201$ & $0.186$ & $-0.580$ & $0.000$  \\ 
\hspace{15pt}Intersections & $0.608$ & $0.004$ & $0.075$ & $-0.159$ & $0.221$  \\ 
\\
Target Density = 15, Subjective Metrics\\ 
\hspace{15pt}Bias & $1.000$ & $1.074$ & $0.253$ & $0.574$ & $1.549$  \\ 
\hspace{15pt}Team (Q1) & $1.422$ & $-0.026$ & $0.256$ & $-0.706$ & $0.463$  \\ 
\hspace{15pt}Competence (Q2) & $1.513$ & $0.065$ & $0.242$ & $-0.444$ & $0.627$  \\ 
\hspace{15pt}Understand Bot Intent (Q3) & $5.216$ & $0.424$ & $0.349$ & $-0.113$ & $1.094$  \\ 
\hspace{15pt}Understand Human Intent (Q4) & $41.769$ & $0.817$ & $0.384$ & $0.000$ & $1.466$  \\ 
\hspace{15pt}Human Contributed More (Q5) & $1.595$ & $0.093$ & $0.212$ & $-0.314$ & $0.616$  \\ 
\hspace{15pt}Easy To Play With (Q6) & $290.561$ & $1.008$ & $0.366$ & $0.333$ & $1.753$  \\ 
\hspace{15pt}Fun To Play With (Q7) & $239769.399$ & $1.720$ & $0.459$ & $0.792$ & $2.580$  \\ 
\hspace{15pt}Similar Playing Style (Q8) & $28.285$ & $0.689$ & $0.334$ & $0.000$ & $1.262$  \\ 

\bottomrule
\end{tabular}
}
\end{table}

\FloatBarrier
\section{Additional Results for Experiment 2}
\label{appendixmoreresultsexp2}
\renewcommand{\thefigure}{F\arabic{figure}}  
\setcounter{figure}{0}  

\begin{figure}[H]
    \centering
    \includegraphics[width=1.0\textwidth]{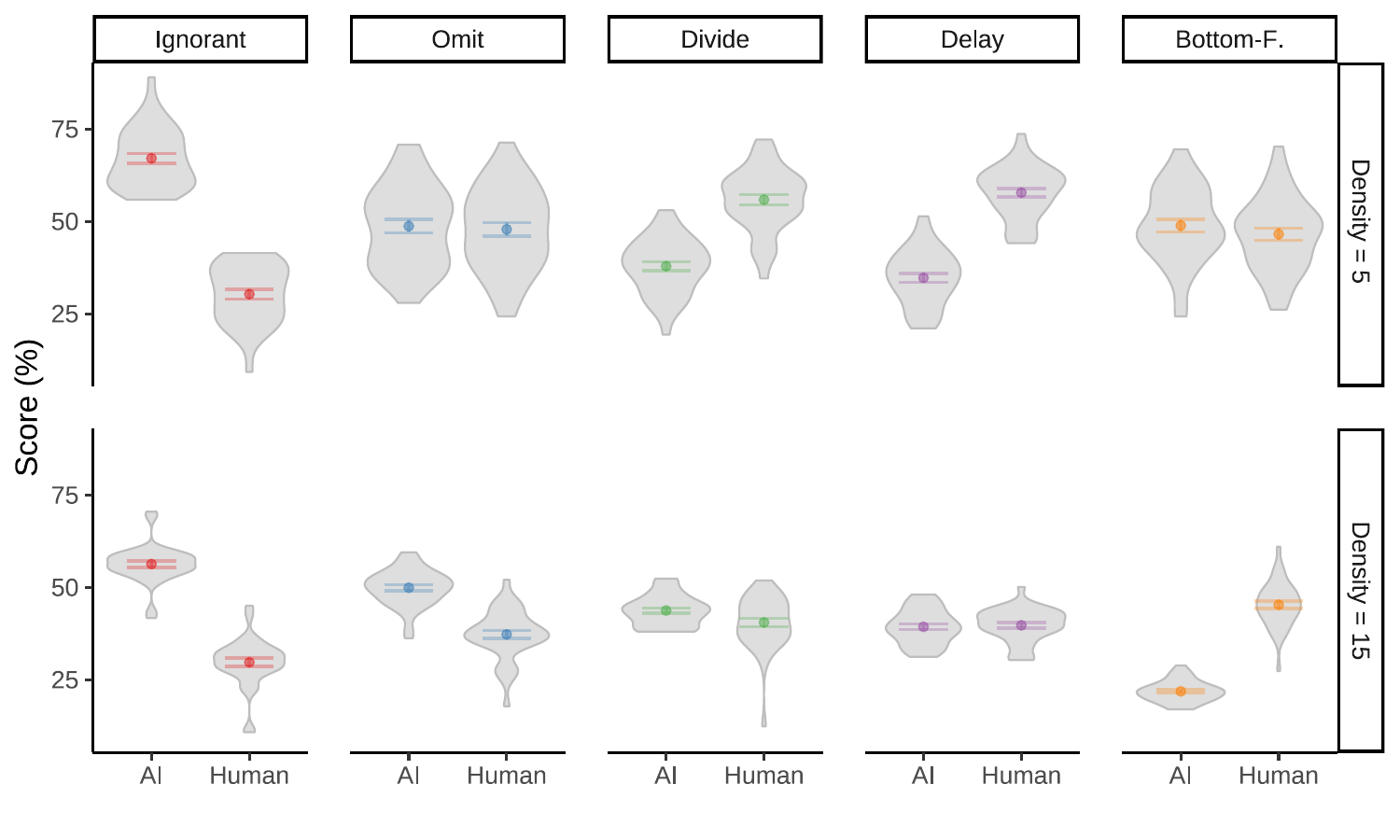}
    \caption{Mean proportional score by agent type, AI type and density condition, with standard error bars for Experiment 2. The gray violins demonstrate the distribution of data points. }
    \label{fig:score_e2}
\end{figure}

\begin{figure}[H]
    \centering
    \includegraphics[width=1.0\textwidth]{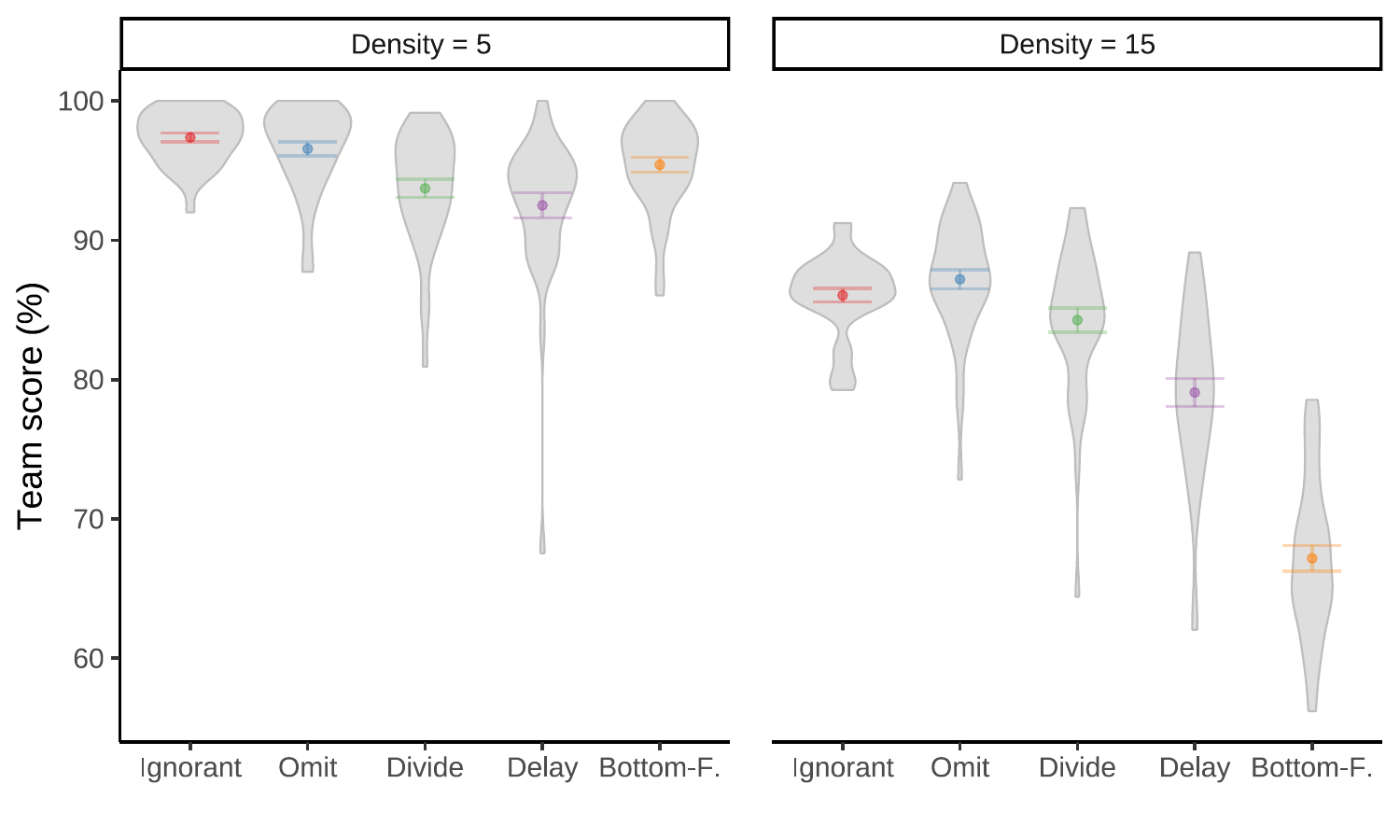}
    \caption{Mean proportional team score (human + AI) by AI type and density condition with standard error bars for Experiment 2. The overlaid gray violins show how data points are distributed.}
    \label{fig:team_e2}
\end{figure}

\begin{figure}[h]
    \centering
    \includegraphics[width=1.0\textwidth]{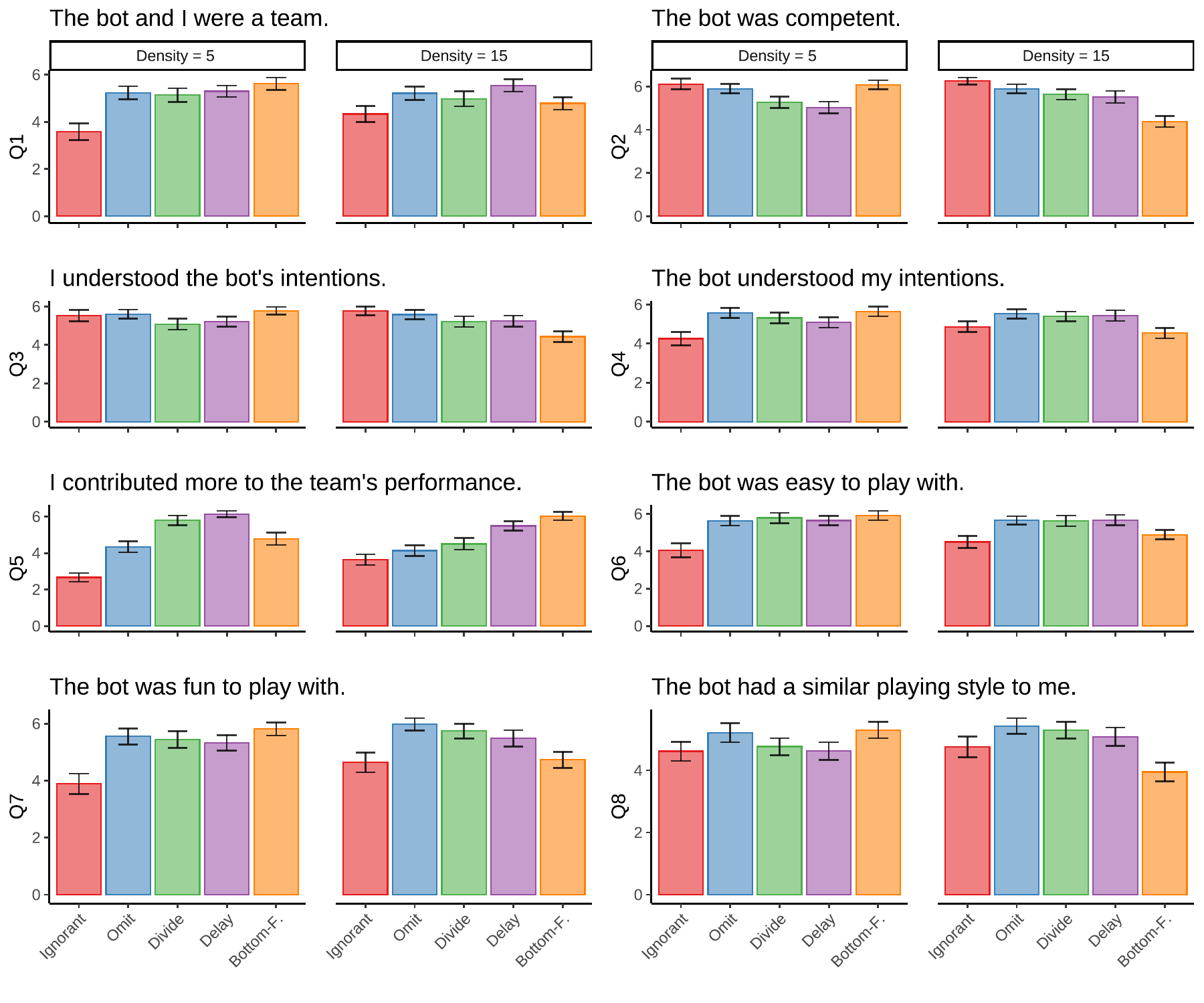}
    \caption{Mean questionnaire scores by AI agent type and target density in Experiment 2. Questions were rated on a 7-point Likert-scale. Error bars indicate the standard error from the mean.}
    \label{fig:survey_e2}
\end{figure}

\begin{figure}[H]
    \centering
    \includegraphics[width=1.0\textwidth]{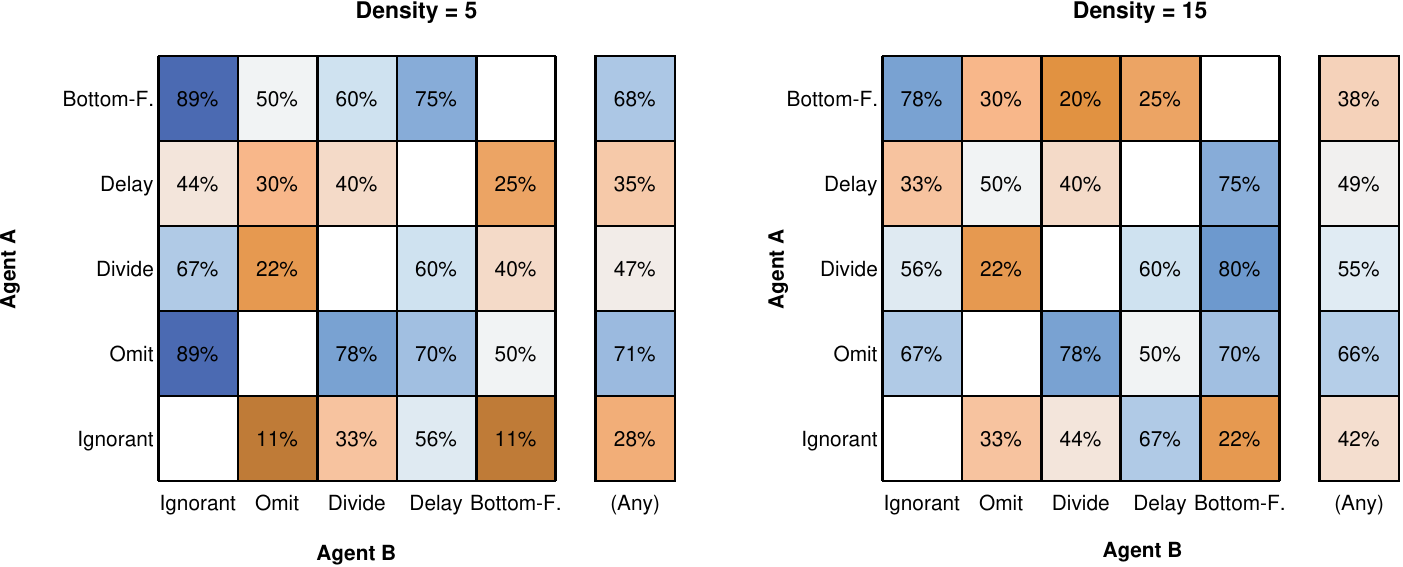}
    \caption{Choice preferences across pairs of agents for each target density condition in Experiment 2. Each matrix cell indicates the percentage of participants preferring the row-associated agent A over the column-associated agent B. The side column shows the overall preference percentage for each row agent across all pairings. Asterisks denote choice percentages that significantly deviate from 50\%, indicated by a Bayes factor greater than 10. Note that results here are averaged across different presentation orders of the agents (e.g., agent A could have been presented first or second in the experiment).}
    \label{fig:choice_e2}
\end{figure}

\end{appendices}


\FloatBarrier
\clearpage


\end{document}